%% file: main.tex
\documentclass[10pt,twocolumn,letterpaper]{article}

\input{preamble}

\input{defs}

\begin{document}
\input{sec/teaser}
\input{sec/0_abstract}    
\input{sec/1_intro}
\input{sec/2_related_work}

\input{sec/3_background}
\input{sec/4_method}

\input{sec/5_optimization}

\input{sec/6_experiments}

\input{sec/7_conclusion}

{
    \small
    \bibliographystyle{ieeenat_fullname}
    \bibliography{main}
}

\input{sec/X_suppl}

\end{document}

%% file: preamble.tex
\usepackage[pagenumbers]{iccv} %

\definecolor{iccvblue}{rgb}{0.21,0.49,0.74}
\usepackage[pagebackref,breaklinks,colorlinks,allcolors=iccvblue]{hyperref}

\def\paperID{} %
\def\confName{CVPR}
\def\confYear{2026}

\title{Geometric Neural Distance Fields for Learning Human Motion Priors}

\author{Zhengdi Yu$^{1}$,  \hspace{1 mm} Simone Foti$^{1}$, \hspace{1 mm}  Linguang Zhang$^{2}$, \hspace{1 mm} Amy Zhao$^{2}$, \\
\hspace{1 mm} Cem Keskin$^{2}$,  \hspace{1 mm} Stefanos Zafeiriou$^{1}$, \hspace{1 mm} Tolga Birdal$^{1}$\\
$^{1}$Imperial College London \qquad $^{2}$ Meta Reality Labs\\
}

\usepackage{url}
\usepackage{enumitem}
\usepackage{color}
\usepackage{graphicx}
\usepackage{amsmath}
\usepackage{mathtools}
\usepackage{amsthm}
\usepackage{caption}
\usepackage{tabularray}
\usepackage{multirow}
\usepackage{comment}
\usepackage{bm}
\usepackage{booktabs}
\usepackage{colortbl}
\usepackage{xcolor}
\usepackage{graphicx}
\usepackage{blkarray}
\usepackage{bigstrut}
\usepackage[dvipsnames]{xcolor}
\usepackage[capitalize]{cleveref}
\usepackage{bm}
\usepackage{comment}
\usepackage{wrapfig}
\usepackage[T1]{fontenc} 

\renewcommand{\paragraph}[1]{{\vspace{1mm}\noindent \bf #1}.}

\newcommand{\name}{{NRMF}}
\newcommand{\RMF}{{RMF}}

\definecolor{steelblue}{HTML}{5E7B8B}
\definecolor{SoftLavender}{HTML}{BFA2D3}
\usepackage{algorithm,algpseudocode}
\algrenewcommand\algorithmicrequire{\textbf{Input:}}
\algrenewcommand\algorithmicensure{\textbf{Output:}}

%% file: defs.tex
\newcommand{\x}{\mathbf{x}}

\newcommand{\Sb}{\mathbf{S}}

\newcommand{\R}{\mathbb{R}}

\newcommand{\G}{\mathbf{G}}
\newcommand{\Gx}{\mathbf{G}_{\mathbf{x}}}
\newcommand{\Id}{\mathbf{I}}

\newcommand{\zero}{\mathbf{0}}

\newcommand{\Rot}{\mathbf{R}}

\newcommand{\hb}{\mathbf{h}}

\newcommand{\X}{\mathbf{X}}
\newcommand{\z}{\mathbf{z}}

\newcommand{\vb}{\mathbf{v}}

\renewcommand{\v}{\mathbf{v}}

\newcommand{\trace}{\mathrm{Tr}}

\newcommand{\bxi}{\bm{\xi}}

\newcommand{\Loss}{\mathcal{L}} %

\newcommand{\Vertices}{\bm{V}}

\newcommand{\pose}[0]{\bm{\theta}}

\newcommand{\roottrans}{\mathbf{t}_{\mathrm{r}}}

\newcommand{\posevel}{\dot{\pose}}
\newcommand{\poseacc}{\ddot{\pose}}
\newcommand{\state}{\X}
\newcommand{\parpose}{\Phi}
\newcommand{\parvel}{\Psi}
\newcommand{\paracc}{\Xi}
\newcommand{\parall}{\Gamma}
\newcommand{\fall}{f_{\parall}}
\newcommand{\fpose}{f_{\parpose}^R}
\newcommand{\fvel}{f_{\parvel}^{\omega}}
\newcommand{\facc}{f_{\paracc}^{\dot{\omega}}}

\newcommand{\data}{\mathcal{D}}
\newcommand{\datapose}{\mathcal{D}_{\theta}}
\newcommand{\datavel}{\dot{\mathcal{D}}_{\theta}}
\newcommand{\dataacc}{\ddot{\mathcal{D}}_{\theta}}

\newcommand{\PoseProj}{\Pi^{R}}
\newcommand{\VelProj}{\Pi^{\omega}}
\newcommand{\AccProj}{\Pi^{\dot{\omega}}}

\newcommand{\obs}{{\mathbf{O}}}
\newcommand{\Lsmooth}{\Loss_{\mathrm{smooth}}}

\newcommand{\Lbl}{\Loss_{\mathrm{bl}}}

\newcommand{\Lshape}{\mathcal{L}_{\shape}}
\newcommand{\Lpose}{\mathcal{L}_{\pose}}
\newcommand{\Lreg}{\mathcal{L}_\mathrm{reg}}
\newcommand{\Lcontact}{\mathcal{L}_\mathrm{contact}}
\newcommand{\Lch}{\mathcal{L}_\mathrm{ch}}

\newcommand{\Ldata}{\mathcal{L}_\mathrm{data}}
\newcommand{\Ltransition}{\mathcal{L}_{\posevel}}
\newcommand{\Lacceleration}{\mathcal{L}_{\poseacc}}

\newcommand{\joints}{\bm{J}}

\newcommand{\shape}{\bm{\beta}}

\newcommand{\bones}{\mathbf{B}}

\newcommand{\joint}{\mathbf{j}}

\newcommand{\curve}{\gamma}
\newcommand{\dcurve}{\dot{\gamma}}

\newcommand{\Amb}{\mathcal{X}}
\newcommand{\Man}{\mathcal{M}}

\newcommand{\TxM}{\T_{\x}\Man}

\newcommand{\T}{\mathcal{T}}

\newcommand{\TR}{\mathcal{T}_{\Rot}}
\newcommand{\grad}[1]{\mathrm{grad}#1}
\newcommand{\SO}{\mathrm{SO}(3)}
\newcommand{\TSO}{\mathcal{T}\mathrm{SO}(3)}
\newcommand{\TSOR}{\mathcal{T}_\mathbf{R}\mathrm{SO}(3)}
\newcommand{\TPose}{\mathcal{T}_{\pose}\mathrm{SO}(3)^K}
\newcommand{\Skew}{\bm{\Omega}}
\DeclareMathOperator{\skewrm}{skew\!}
\DeclareMathOperator{\sym}{sym\!}
\newcommand{\rod}{\bm{\omega}}
\newcommand{\drod}{\dot{\bm{\omega}}}
\newcommand{\solie}{\mathfrak{so}(3)}

\newcommand{\Exp}{\mathrm{{Exp}}}
\newcommand{\Log}{\mathrm{{Log}}}

\newcommand*\diff{\mathop{}\!\mathrm{d}}

\newcommand{\dR}{\dot{R}}
\newcommand{\dRot}{\dot{\Rot}}

\newcommand{\angvel}{{{\Omega}}}
\newcommand{\angvelvec}{{\bm{\omega}}}
\newcommand{\angvelmat}{{\bm{\Omega}}}

\newcommand{\sota}{SotA}

\crefname{equation}{eq.}{eq.}
\Crefname{equation}{Eq.}{Eq.}
\crefname{theorem}{thm.}{thms.}
\Crefname{Theorem}{Thm.}{Thms.}
\crefname{conjecture}{conj.}{conjs.}
\Crefname{Conjecture}{Conj.}{Conjs.}
\crefname{proposition}{prop.}{props.}
\Crefname{proposition}{Prop.}{Props.}
\crefname{definition}{dfn.}{dfn.}
\Crefname{definition}{Dfn.}{Dfn.}
\crefname{remark}{remark}{remark}
\Crefname{Remark}{Remark}{Remark}
\Crefname{algorithm}{Alg.}{Alg.}

\newtheorem{remark}{Remark}

\newtheorem{prop}{Proposition}
\newtheorem{dfn}{Definition}

\crefname{section}{Sec.}{Secs.}
\Crefname{section}{Sec.}{Secs.}
\crefname{equation}{Eq.}{Eqs.}
\Crefname{equation}{Eq.}{Eqs.}
\crefname{figure}{Fig.}{Figs.}
\Crefname{figure}{Fig.}{Figs.}
\crefname{table}{Tab.}{Tabs.}
\Crefname{table}{Tab.}{Tabs.}
\crefname{thm}{Thm.}{Thms.}
\Crefname{thm}{Thm.}{Thms.}
\crefname{conj}{Conj.}{Conjs.}
\Crefname{conj}{Conj.}{Conjs.}
\crefname{dfn}{Dfn.}{Dfns.}
\crefname{dfn}{Dfn.}{Dfns.}
\crefname{remark}{remark}{remarks}
\Crefname{Remark}{Remark}{Remarks}
\crefname{prop}{Prop.}{Prop.}
\Crefname{prop}{Prop.}{Prop.}
\Crefname{algorithm}{Alg.}{Alg.}
\crefname{appendix}{App.}{apps.}
\Crefname{appendix}{App.}{Apps.}
\crefname{appsec}{appendix}{appendices}
\Crefname{appsec}{Appendix}{Appendices}

\DeclareMathOperator*{\argmin}{arg\min}

\newcommand\norm[1]{\lVert#1\rVert}

%% file: sec/teaser.tex
\twocolumn[{%
\renewcommand\twocolumn[1][]{#1}%
\vspace{-5mm}
\maketitle
\vspace{-5mm}
\begin{center}
    \centering\vspace{-5mm}
    \captionsetup{type=figure}
    \includegraphics[width=\textwidth]{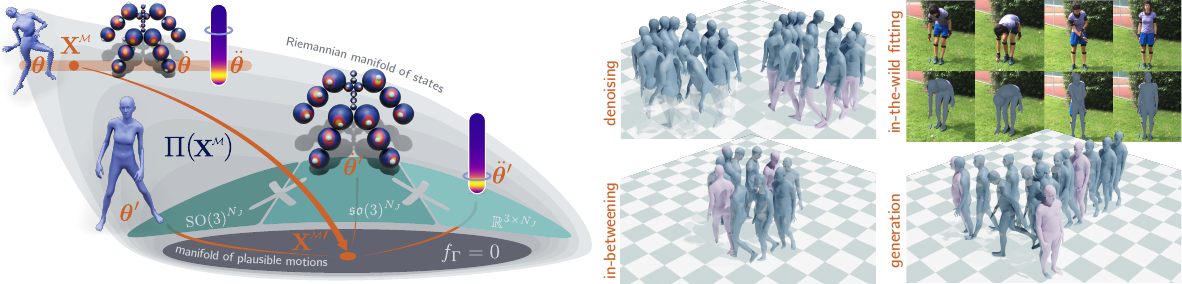}
    \captionof{figure}{
    \textbf{\name} is a general-purpose, expressive and robust unconditional motion prior. It models the space of plausible \textbf{poses} ($\pose$), \textbf{transitions} ($\posevel$) and \textbf{accelerations} ($\poseacc$) on the zero-level set of a \textbf{geometric neural distance field}. This implicitly captures the data distribution. 
    Poses are depicted alongside their transitions and accelerations, which are visualized as blue dots onto the per-joint distributions of learned transitions and as blue rings around the magnitude distribution of all accelerations.
    We develop projection ($\Pi$) and integration algorithms to deploy \name~into several applications as shown on the right. %
    }
    \label{fig:teaser}
\end{center}%
}]

%% file: sec/0_abstract.tex
\begin{abstract}
We introduce Neural Riemannian Motion Fields (\name), a novel 3D generative human motion prior that enables robust, temporally consistent, and physically plausible 3D motion recovery. Unlike existing VAE or diffusion-based methods, our higher-order motion prior explicitly models the human motion in the zero level set of a collection of neural distance fields (NDFs) corresponding to pose, transition (velocity), and acceleration dynamics. Our framework is rigorous in the sense that our NDFs are constructed on the product space of joint rotations, their angular velocities, and angular accelerations, respecting the geometry of the underlying articulations. We further introduce: (i) a novel adaptive-step hybrid algorithm for projecting onto the set of plausible motions, and (ii) a novel geometric integrator to “roll out” realistic motion trajectories during test-time-optimization and generation. 
Our experiments show significant and consistent gains: trained on the AMASS dataset, \name~remarkably generalizes across multiple input modalities and to diverse tasks ranging from denoising to motion in-betweening and fitting to partial 2D / 3D observations. Our project page is available \href{https://circle-group.github.io/research/NRMF/}{\textbf{here}}.
\vspace{-3mm}
\end{abstract}

%% file: sec/1_intro.tex
\section{Introduction}
\label{sec:intro}
Motion is intrinsic to human experience. As humans, we are in perpetual motion, engaging with our surroundings in rich and diverse ways. Accurately recovering 3D human motion from sparse or noisy data is therefore not only a fundamental challenge in computer vision, but also a critical enabler for a myriad of applications, from immersive virtual reality and animation to clinical diagnostics, and sports analysis. Despite remarkable progress in learning-based techniques and model-based reconstructions, current methods still struggle to deliver reliable, temporally consistent, and physically plausible 3D motion recovery while being able to generate diverse and \emph{lifelike} motions.
The challenges stem from the intrinsic complexity of human motion, compounded by occlusions, rapid movements, and inherent ambiguities in 2D observations.

To address these challenges, we focus on the problem of building \emph{general-purpose}, \emph{expressive} and \emph{robust} models that capture the full dynamics of \emph{realistic} human motion.
A conventional way to handle sequences of articulations is to model static poses, frame-by-frame as done in NRDF~\cite{he24nrdf}, PoseNDF~\cite{tiwari22posendf} or DPoser~\cite{lu2023dposer}. Yet, processing frames independently jeopardizes generalization and temporal consistency. 
As a remedy, modern models of motion have shifted toward dynamics-based methods, many of which rely either on 
VAEs~\cite{shi2023phasemp,rempe2021humor} suffering from \emph{posterior-collapse} and increased accumulation of errors over time (\emph{drift}) or leverage the recent diffusion modeling~\cite{tevet2023human, shafir2024human}, which while promising, tend to capture short pose sequences in clean settings or require additional conditioning such as action or textual descriptions~\cite{petrovich2023tmr,azadi2023make}. Diffusion-based models further suffer from increased inference times and their conditioning to observations (\emph{inversion}) is far from being trivial. While RoHM~\cite{zhang2024rohm} tries to bridge this gap, as we will demonstrate, it suffers from low estimation accuracy, due to the limited handling of higher-order motion dynamics, such as acceleration, causing oversmoothed transitions and drift. MoManifold~\cite{Dang2024MoManifold} factors in the accelerations. Yet, beyond being limited to short, fixed time windows, its \emph{decoupled manifold} treats each joint’s acceleration in isolation, neglecting the essential inter-dependencies of the body joints. 

In this work, we introduce \textbf{Neural Riemannian Motion Fields} (\textbf{\name}), a novel human motion prior that models the manifold of plausible motions as a collection of three neural distance fields (NDFs), each corresponding to 0th-order (pose), 1st-order (pose transitions), and crucially, 2nd-order (acceleration) dynamics, captured as the zero level sets of dedicated conditioned, implicit neural networks (\cref{fig:pipeline}b). 
We rigorously respect the underlying geometry of articulated motion by operating in the product space of joint rotations, angular velocities, and angular accelerations. 
To map an arbitrary articulated motion onto a plausible one (\cref{fig:pipeline}a), we introduce a three-stage adaptive-step hybrid gradient descent. We further devise a novel \emph{geometric} and \emph{robust integrator} to \emph{rollout} physically realistic motion sequences, while correcting errors in all dynamic components. 
Together, these enable a plethora of applications as exemplified in~\cref{fig:teaser}.

After training on the large AMASS~\cite{AMASS:2019} motion capture dataset, we extensively evaluate \name~on numerous tasks including motion estimation, denoising, in-betweening, and fitting to various input modalities, over various datasets including AMASS, i3DB~\cite{monszpart2019iMapper}, 3DPW~\cite{vonMarcard2018}, EgoBody~\cite{zhang2022egobody} and PROX~\cite{hassan2019prox}. Our quantitative results show \textbf{significant and consistent improvements} across several standard metrics for assessing human motion. Notably, a qualitative investigation into the learned pose, transition and acceleration distributions reveals the expressivity of our method.
To summarize, our contributions are:
\begin{enumerate}[noitemsep]
\item We propose to model poses, transitions and accelerations as the zero level sets of corresponding product NDFs.
\item We introduce new algorithms for projecting onto the zero level sets as well as for rolling out physically plausible motion sequences.
\item We ensure that all our algorithms respect the geometry of articulated rotations. 
\item From sampling motions to fitting 2D/3D observations, we demonstrate that \name~surpasses all prominent state-of-the-art methods across several tasks and metrics.
\end{enumerate} 
We additionally present a condensed primer on the geometry of articulated motions in our suppl. material as a self-contained background. We will make our implementation publicly available upon publication.

%% file: sec/2_related_work.tex
\section{Related Work}\label{sec:related_work}
\vspace{-2mm}
\paragraph{Unconditional human pose and motion priors}
Human bodies have been modeled by unconditional priors of various different kinds including Gaussian Processes~\cite{yan2019convolutional}, VAEs~\cite{pavlakos2019expressive}, NDFs~\cite{he24nrdf,tiwari22posendf,chibane2020neural}, diffusion models~\cite{lu2023dposer,ci2023gfpose}. 
Yet, per-frame inference is not sufficient to capture the full complexity of motion. To this end, various works such as MotionVAE~\cite{ling2020MVAE}, HuMoR~\cite{rempe2021humor} and 
PhaseMP~\cite{shi2023phasemp} leveraged VAEs in an autoregressive fashion. With the proliferation of diffusion models and transformers, modern approaches such as~\cite{tevet2023human,shafir2024human,huang2024stablemofusion,yu2024towards,kulkarni2024nifty} leverage transformer-based diffusion frameworks to generate diverse motion. Unlike these works, we take a fundamentally different approach and offer a strong generative prior that is deployable for recovering human motion from observations.

\paragraph{Monocular human pose and motion recovery} 
Approaches for estimating human pose and shape from monocular observations can be categorized into regression-based and optimization-based. Regression-based methods predict pose directly from images or videos using deep networks \cite{TRACE,kocabas2021pare,cai2023smplerx,goel2023humans,zhang2023pymaf,xu2023smpler,baradel2022posebert} which are typically based on the parameteric models of SMPL \cite{loper2015smpl}, SMPL-H \cite{smplh:SIGGRAPHASIA:2017} or SMPL-X \cite{pavlakos2019expressive}, but often lack temporal consistency, and fail under occlusion. Optimization-based approaches \cite{rempe2021humor,shi2023phasemp,zhang2024rohm} fit parametric models to observations while enforcing temporal priors, achieving better robustness but at the cost of high computation, and sensitivity to initialization. Recent methods also integrate learned motion priors via diffusion \cite{tevet2023human,shafir2023human}, improving plausibility and infilling. RoHM \cite{zhang2024rohm} reconstructs motion via diffusion denoising of root and local pose spaces with image and physics guidance. However, it still suffers from oversmoothing and cannot fully exploit higher-order dynamics. Our method NRMF explicitly models pose, velocity, and acceleration via neural distance fields on Riemannian motion manifolds, enabling physically plausible recovery across diverse inputs.

%% file: sec/4_method.tex
\vspace{-1mm}\section{Neural Riemannian Motion Fields}\label{sec:method}
\vspace{-1mm}
We start by explaining \textbf{Riemannian Motion Fields} to model realistic articulated motions in~\cref{sec:model} and introduce our novel projection algorithm to map onto this space while adhering to the manifold of joint rotations, transitions and accelerations. We then propose \name~and a novel method for sampling articulated motions, to generate desired training data, in~\cref{sec:NRMF}. Before delving into method specifics, we briefly review the $\SO$-kinematics and provide a detailed, self-contained exposition in our suppl. material.

\paragraph{Background}
We are particularly interested in the non-commutative Lie group of finite rigid body rotations:
\begin{dfn}[$\SO$]
    The set of rotation matrices admit a group structure under matrix multiplication:
    \begin{equation}
\SO=\left\{\Rot \in \R^{3 \times 3}: \Rot^{\top} \Rot=\mathbf{I}, \operatorname{det}(\Rot)=1\right\}.
\end{equation}
\end{dfn}
We now briefly characterize the properties of $\SO$.
\begin{prop}
The tangent space of $\SO$ at $\Rot$ is defined throughout the paper as the \emph{left-invariant vector fields}:
\begin{equation}
    \label{eq:tso3}
    \TSOR=\left\{\Rot\Skew \,:\, \Skew\in\solie\right\}, 
\end{equation}
where $\solie$ is given by the set of \emph{skew-symmetric} tensors:
\begin{equation}
    \solie=\left\{\Skew \in \R^{3 \times 3}: \Skew^{\top}=-\Skew\right\}. 
\end{equation}
\end{prop}
\begin{proof}
Differentiating the constraints that specify $\SO$ leads us to the definition of skew-symmetric matrices:
\begin{align}
    \label{eq:proof-skewsym}
        \frac{\diff(\Rot^\top\Rot)}{\diff t} = \dRot^\top\Rot + \Rot^\top \dRot  = 0 \implies \Skew = -\Skew^\top, %
    \end{align}
    where $\Skew=\Rot^\top\dRot$ is a skew symmetric matrix and the vector $\dRot=\Rot\Skew$ is tangent to $\Rot$. 
\end{proof}
We also need the notions of angular velocity and acceleration on $\SO$ to accurately model articulated motions.
\begin{dfn}[Angular velocity]
    The \emph{angular velocity} of the geodesic motion on $\SO$ is represented as a time ($t$) dependent curve $\angvelvec_t:=\omega(t):=\skewrm\,(\angvel(t))$ extracted from:
    \begin{equation}\label{eq:angvel}
        \angvelmat_t = \Rot_t^{-1}\dRot_t,
    \end{equation}
    where $\skewrm(\cdot)$ transforms the matrix into a \emph{skew-vector} and 
    the \textbf{angular velocity matrix} $\angvelmat_t:=\Omega(t)$ is skew-symmetric.
\end{dfn}
\begin{prop}[Angular acceleration]
The time-derivative of angular velocity defines the angular acceleration:
\begin{equation}\label{eq:angacc}
    \frac{\diff [\angvelvec_t]_x}{\diff t}:=[\drod_t]_x:=\ddot{\Rot}_t=\Rot_t\left(\angvelmat_t^2+\dot{\angvelmat_t}\right),
\end{equation}
where $\angvelmat_t^2+\dot{\angvelmat_t}$ is also skew-symmetric and $\dRot_t:={\diff \Rot_t/}{\diff t}$.
\end{prop}
\noindent %
In practice, we use a central differencing scheme to estimate the angular velocity and acceleration. As we show in our suppl. material, this is a good approximation of the true acceleration computed via $\log$-maps and \emph{parallel transport}. %

Our work differs significantly from modeling motion in pure $SO(3)$~\cite{bastian2025forecasting,haarbach2018survey} also because of the specific \emph{product manifold} structure which we discuss next.

\subsection{Modeling of Plausible Articulated Motions}
\label{sec:model}
We consider an input motion sequence $\{\state_t\}_{t=0}^T$ composed of individual states $\state_t$:
\begin{dfn}[State]
    We represent the state of a moving articulated body as a time dependent function $x(t):t\to\state_t$ producing the matrix $\state_t=x(t)$ composed of a root translation $\roottrans \in \R^3$, body joint angles $\pose \in \SO^{N_{J}}$, their velocities $\posevel \in \TSO^{N_{J}}$ and angular accelerations $\poseacc \in \R^{3\times {N_{J}}}$:
\begin{align}
\label{eqn:staterep}
    \state = [ \;\; \roottrans \;\; \pose \;\; \posevel \;\; \poseacc \;\;]\;\in\;(\R^3\times\Man),
\end{align}
where $N_J$ refers to the total number of joints excluding the hand joints ($N_{J}=22$ for SMPL model) and $\Man$ denotes the product space of the state vector including the root orientation: $\Man=\left(\SO^{N_J}\times\solie^{N_J}\times \R^{3 \times N_J}\right)$.
\end{dfn}
The first part of the state involving root translation is known as the \textbf{global motion}, whereas the rest parameterize the \textbf{local motion}, the motion of the skeleton.
Part of the state, $(\roottrans,\pose)$, parameterizes the articulated object model (e.g., SMPL for humans~\cite{loper2015smpl,smplh:SIGGRAPHASIA:2017} or MANO for hands~\cite{Romero2017Embodied}) which is a differentiable function $M(\roottrans,\pose,\shape)$ %
mapping joints $\joints \in \R^{3\times J}$ and shape parameters $\shape \in \R^{N_{\beta}}$ to the vertices $\Vertices \in \R^{3\times N_{\Vertices}}$ of an articulated object, where $N_{\Vertices}=6890$ for SMPL human bodies.

Instead of parameterizing the entire motion sequence as done in~\cite{he2022nemf}, we propose to model the current state (pose, transition and acceleration) while respecting the geometry:
\begin{prop}[Riemannian Motion Fields (RMFs)]
    Given the state, we model the manifold of instantaneous \textbf{realistic and plausible motions} of an articulated body as the zero level set of an implicit function $\fall: \Man \to \R_+^3$:
    \begin{equation}
\mathcal{S} = \{\state^{\Man} \in \Man \,\mid\, \fall(\state^{\Man}) = \zero\} \textrm{,}
\end{equation}
such that the range of $\fall$ (parameterized by $\parall$) represents the unsigned geodesic distance to the closest plausible pose on the manifold. To account for the interdependence of pose, velocity and acceleration we disentangle $\fall$ into conditional components as:
\begin{equation}
  \fall = 
  \left[
  \begin{array}{c}
    \fpose({\pose}) \\
    \fvel({\posevel\,\mid\, \pose}) \\
    \facc({\poseacc\,\mid\, \pose, \posevel}) \\
  \end{array}
  \right]
  \begin{array}{l}
    \emph{(pose~field)} \\
    \emph{(transition~field)} \\
    \emph{(acceleration~field)} \\
  \end{array}
  \tag{RMF}
\end{equation}
where $\fpose:\SO^{N_J}\to\R_+, \fvel:\solie^{N_J}\to\R_+$ and $\facc:\R^{3 \times N_J}\to\R_+$ account for the plausibility in pose, velocity and acceleration, respectively. $\parall=[\parpose, \parvel, \paracc]$. 
\end{prop}
\begin{figure}[t]
        \centering
        \includegraphics[width=\columnwidth]{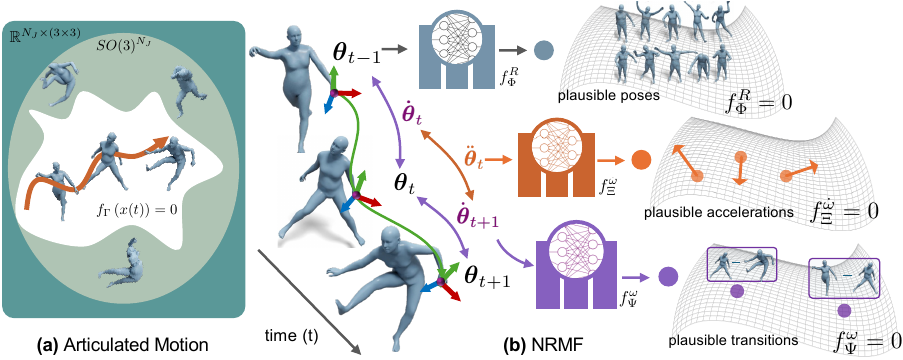}
        \caption{Neural Riemannian Motion Fields (\textbf{\name}) models the motion in the zero-level-set of three disjoint distance fields.\vspace{-4mm}}
        \label{fig:pipeline} 
\end{figure}
We are now ready to state our iterations for projecting onto this zero level set specified by $\fall$:
\begin{prop}[\RMF-Grad]
    A geometry-aware single step projection of a given state onto our motion manifold involves the updates:
    \begin{align} 
    \pose_{t+1}:=\PoseProj(\pose_t) &= \mathrm{Exp}_{\pose_t}\left(-\alpha_{\theta} \fpose(\pose_t)\,\frac{\grad{\fpose(\pose_t)}}{\|\grad{\fpose(\pose_t)}\|}\right)\nonumber\\
    \posevel_{t+1}:=\VelProj_t(\posevel_t)  &= \posevel_t-\alpha_{\dot{\theta}} \fvel(\posevel_t)\,\frac{\nabla{\fvel(\posevel_t\,\mid\, \pose_t)}}{\|\nabla{\fvel(\posevel_t\,\mid\, \pose_t)}\|}\\
    \poseacc_{t+1}:=\AccProj_t(\poseacc_t)  &= \poseacc_t-\alpha_{\dot{\omega}} \facc(\poseacc_t)\,\frac{\nabla{\facc(\poseacc_t\,\mid\, \pose_t, \posevel_t)}}{\|\nabla{\facc(\poseacc_t\,\mid\, \pose_t, \posevel_t)}\|}
    \textnormal{,}\nonumber
    \end{align}
    where $\alpha_*$ corresponds to individual learning rates and $\grad$ and $\Exp$ denote the \textbf{Riemannian gradient} and \textbf{Exponential maps} as we provide in our suppl. material.
\end{prop}
\begin{remark}
    Note that under ideal conditions where transition (velocities) are obtained computationally, we expect:
    \begin{equation}
        \VelProj\left( \posevel \right) \approx \left(\PoseProj\left(\pose\right)^\top \dot{\PoseProj\left(\pose\right)} \right),
    \end{equation}
    and leverage the right hand side, accounting for the use of noisy input velocities. Likewise, $\AccProj( \poseacc ) \approx {\diff (\VelProj(\posevel))}/{\diff t} $.
\end{remark}

\paragraph{A geometry-aware robust integrator}
We now provide a novel \emph{deterministic} integration algorithm for \textbf{rolling out} motion sequences starting from a plausible initial state. Equivalently, it will allow us to \emph{rebuild} a consistent state sequence from the projected measurements, a key step in motion generation and test-time-optimization as we will explain later.
\begin{prop}[\RMF-Integrator]
    For a moving articulated body $x(t):t\to\state_t$, the ``ideal'' continuous dynamics are governed by the differential equations in~\cref{eq:angvel,eq:angacc}. We then apply a geometric, projected \textbf{Euler integration} to those factoring in our priors. Given a sequence of plausible accelerations, this leads to the following discrete updates:
    \begin{align} 
            \posevel_{t+1} &= \VelProj(\posevel_{t}+\lambda_t\poseacc_{t})\\
            \pose_{t+1} &= \PoseProj\left(\Exp_{\pose_{t}}\bigl(\alpha_t\, [{\posevel}_{t}]_x\bigr)\right).
    \end{align}
    In practice, we first update the velocities before integrating for the rotations.
\end{prop}
The pseudocode for our integrator is provided in our suppl. material.

\begin{remark}[Relation to known operators and integrators]
    $\PoseProj(\pose_t)$ corresponds to NRDF~\cite{he24nrdf} while the other two projectors resemble Euclidean PoseNDF~\cite{tiwari22posendf} updates but for velocities and accelerations. Beyond ensuring plausibility of motion, our \RMF-Integrator can be thought of as a \textbf{denoising} of the trajectory and its derivatives or a \textbf{drift correction} that helps mitigate the accumulation of errors. 
\end{remark}

\subsection{Learning Riemannian Motion Fields (\name)}
\label{sec:NRMF}
We now describe how we construct $\mathcal{S}$, \ie, learn all components of $f_{\Gamma}$ indicating distances to the zero level set.
\begin{prop}[\name]
    As shown in \cref{fig:pipeline}, we model $\fall$ hierarchically by three individual neural fields corresponding to different orders of derivatives in the equation of motion, trained to predict the distance to the closest example from dataset $\data=[\datapose,\datavel,\dataacc]$ as follows\footnote{Note, for brevity, we dropped the conditionals.}: 
\begin{equation}
\label{eq:training}
\resizebox{\linewidth}{!}{$
{\parall^\star} = 
\left[
\begin{aligned}
    {\parpose^\star}&=\argmin_{\parpose} \sum\limits_{i=1}^N \norm{ \fpose(\pose_i) - \min_{\pose^\prime\in\datapose} d_{\SO}^{N_J}(\pose_i, \pose^\prime)}\\
{\parvel^\star}&=\argmin_{\parvel} \sum\limits_{i=1}^N \norm{ \fvel(\posevel_i) - \min_{\posevel^\prime\in\datavel} d_{\solie}^{N_J}(\posevel_i, \posevel^\prime)}\\
{\paracc^\star}&=\argmin_{\paracc} \sum\limits_{i=1}^N \norm{ \facc(\poseacc_i) - \min_{\poseacc^\prime\in\dataacc} d_{\R^{3}}^{N_J}(\poseacc_i, \poseacc^\prime)}
\end{aligned}
\right]\textnormal{.}\nonumber$}
\end{equation}
We call $\fall^\star$, learned in this way, a \emph{neural Riemannian motion field}.
\end{prop}
Each individual function is modeled by a combination of a hierarchical network and an MLP decoder, similar to NRDF~\cite{he24nrdf} and Pose-NDF~\cite{tiwari22posendf}. 
We obtain $\grad \fall^\star$ via backprop, additionally followed by \emph{egrad2rgrad} for the poses themselves. 
Given a dataset $\datapose = \{\pose_i\}_{1\leq i \leq N}$ of articulated poses, we obtain the dataset of transitions ($\datavel$) and accelerations ($\dataacc$) by computing the angular velocities and accelerations as explained earlier. 

\begin{remark}
The first part of this equation corresponds to the recent NRDF~\cite{he24nrdf}, which builds upon PoseNDF~\cite{tiwari22posendf} and correctly treats the Riemannian geometry of articulated 3D poses using quaternions. The acceleration field in the last part vaguely resembles MoManifold~\cite{Dang2024MoManifold}, while properly using the joints orientations instead of 3D positions. 
The \textbf{transition prior} in the middle row bears similarities with various works which model transitions~\cite{shi2023phasemp,rempe2021humor}, inspired mostly by \cite{rempe2021humor}. A majority of these works use some form of autoencoding, which is fundamentally different from our approach.
To the best of our knowledge, we are the firsts to properly model all components of the kinematics, while respecting the geometry of parameters.
\end{remark}

%% file: sec/5_optimization.tex
\section{NRMF as Versatile Generative Priors}
In this section, we introduce multiple downstream applications. Utilizing our NRFM as a generative prior, we can enable: 1) test-time optimization, 2) motion generation and 3) motion in-betweening, as shown in \cref{fig:teaser}.

\paragraph{Test-time optimization}
We introduce a novel runtime optimization objective to recover a plausible (\emph{likely} under the NRMF) motion sequence $\{\state_t\}_{t=0}^T$ conditioned on a given sequence of 2D or 3D data observations, \eg joint detections, point clouds, described as $\obs_{0:T}$.
Our mutlti-stage optimization starts from a proper initialization using the pose prior, by minimizing a HuMoR-like objective in stage I:
\begin{equation}
    E_{I}(\roottrans,\pose,\shape) = \Ldata + \lambda_{\shape}\Lshape + \lambda_{\pose}\Lpose + \lambda_{reg}\Lreg
\end{equation}
where $\Lshape=\|\shape\|^{2}$ is the shape prior\footnote{We will later optimize for the shape parameters ($\beta$) per sequence.}. $\Lpose=\fpose(\pose_i)$ is our pose prior term predicting the distance to the nearest plausible pose\footnote{Note that for VPoser~\cite{pavlakos2019smplx} $\Lpose=\|\z\|^{2}_{2}$ where $z$ is the latent code.}. We further employ additional regularization terms $\Lreg$ to constrain the bone lengths and smoothness of the skeleton as described in~\cite{rempe2021humor}. 
Finally, $\Ldata$ denotes the task-dependent data terms and will be different when given different types of inputs: 2D, 3D joints or 3D point clouds. We provide complete details of these in our supplementary material.

\begin{figure*}[t]
    \centering
    \includegraphics[width=\linewidth]{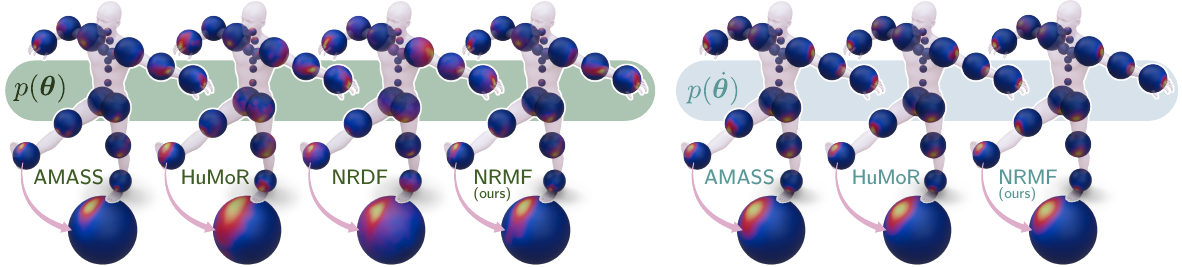}
    \caption{\textbf{Comparison of pose and transition distributions.} Each distribution is estimated from the points obtained intersecting rotated skeletal bones with a sphere centered at the joint. The pose distributions $p(\pose)$ are obtained applying $\pose$ rotations (\textit{left}), while the transition distributions $p(\posevel)$ applying $\posevel$ (\textit{right}). Each sphere effectively displays the range of motion of eah joint.   \vspace{-4mm}}
    \label{fig:all_distribs} 
\end{figure*}
While optimizing $E_{I}$ yields refined estimates $\state$, it cannot handle motion ambiguities since the pose prior only captures $0^{th}$-order distribution. After initialization, we factor in our transition and acceleration priors as $\Ltransition$ and $\Lacceleration$:
\begin{equation}
    \Ltransition := \fvel(\posevel_i) \quad \Lacceleration := \facc(\poseacc_i)
\end{equation}
By further combining foot-floor contact physics constraint terms into $\Lreg$ in stage II, the full objective function reads:
\begin{align}
    E_{II}^\prime(\roottrans,\pose,\shape) = \Ldata 
    &+ \lambda_{\shape}\Lshape 
    + \lambda_{\pose}\Lpose \\
    &+ \lambda_\mathrm{reg}\Lreg 
    + \lambda_{\posevel}\Ltransition 
    + \lambda_{\poseacc}\Lacceleration\nonumber
\end{align}
Moreover, we apply the rollout within each iteration to further increase precision. We provide more details in the supplementary material.

\paragraph{Motion generation \& in-betweening} 
Our motion prior can be utilized to generate plausible and diverse motions. To this end, inspired by \cite{rempe2021humor,Dang2024MoManifold}, we start from a random initialization of all motion components. In particular, given an initial motion state $\state_{0}$, we randomly sample a noisy transition $\posevel_{0}$ and generate the next $\pose_{1}$ using the transition. By repeating this procedure in a rollout manner, we obtain a noisy partial motion sequence with $\{\pose_{t}, \posevel_{t} \}^{T}_{t=0}$. The noisy acceleration $\poseacc$ is then computed based on it to compose the complete initial noisy motion sequences $\{\Tilde{\state_{t}}\}^{T}_{t=0}$. Next, we run our RMF-Integrator taken the sequence as input to denoise and obtain the $\{\state_t\}^{T}_{t=0}$, which is subsequently treated as the initial noisy observation for test-time optimization (Sec. \textcolor{red}{4}) similar to \textbf{MoManifold} to obtain the final clean motion from the randomly composed disordered starting point. For in-betweening, given a sequence of incomplete observations, we preprocess the unobserved poses with \cite{he2022nemf,shoemake1985slerp} and then apply the optimization process with $E_{I}$ nd $E_{I}^\prime$ treating the observed frames as noisy data observation inspired by \cite{yu2025dynhamr}. Unlike \cite{tevet2023human}, our method is not restricted to clean input motions, it can be applied to partial masked body infilling, and it also deduces natural transitions between masked between-frame with $\facc(\posevel_i)$, while respecting the $2^{\text{nd}}$-order dynamics via the acceleration prior $\facc(\poseacc_i)$.

%% file: sec/6_experiments.tex
\section{Experiments}\label{sec:exp}

We now assess the performance of our method across various downstream tasks: (i) motion estimation from 3D observations, (ii) motion refinement for human mesh recovery, (iii) motion estimation from RGB(-D) in-the-wild (2D) observations and (iv) motion generation and in-betweening. Additionally, we conduct ablation studies to demonstrate the contribution of each prior. We recommend viewing the \textcolor{magenta}{supplementary videos} for qualitative results illustrating our approach. Further details of the dataset and experimental setup are provided in our suppl. material.

\subsection{Setup}
\paragraph{Datasets} In this work, we leverage the following datasets:
\begin{enumerate}[noitemsep,topsep=0em,leftmargin=*]
    \item \textbf{AMASS}~\cite{AMASS:2019} is a large-scale human motion database comprising of high-quality and diverse motion captures. For training and evaluation, we subsample the dataset at a rate of 30 Hz and follow the split used in \cite{he24nrdf,tiwari22posendf,pavlakos2019expressive}. We extract transitions (velocities) and corresponding accelerations by sampling 80\% of each motion sequence.
    \item \textbf{3DPW}~\cite{vonMarcard2018} is a popular benchmark for 3D human pose and shape estimation in challenging, real-world settings. Provides high-quality 3D pose annotations captured through synchronized IMUs and motion capture systems, along with RGB videos recorded in diverse outdoor environments. %
    \item \textbf{i3DB}~\cite{vonMarcard2018} is an RGB dataset of human-scene interactions, providing 3D joint annotations along with 3D scene reconstructions. %
    \item \textbf{EgoBody}~\cite{zhang2022egobody} is a large-scale dataset of 3D human motion captures during social interactions in 3D environments, with one subject wearing a head-mounted device recording egocentric multi-modal data, including eye gaze tracking, synchronized multi-view RGBD video, and full-body 3D motion reconstruction. 
    \item \textbf{PROX}~\cite{hassan2019prox} is an RGB-D dataset that provides 3D human pose and shape data in 3D real-world indoor environments with interactions between humans and surroundings. We pre-process them to obtain the observations in the same way as i3DB. %
\end{enumerate}

\begin{table}[t]
\centering
\caption{{Motion and shape estimation from noisy 3D observations (\textbf{motion denoising}).}\vspace{-2mm}}
\label{tab:denoising}
\setlength{\tabcolsep}{4pt}
\resizebox{\columnwidth}{!}{%
\begin{tabular}{lccccccc}
\toprule
\textbf{Method} & \multicolumn{3}{c}{\textbf{Pos. Err. (mm)}} & \multicolumn{4}{c}{\textbf{Motion Prop.}} \\
\cmidrule(lr){2-4} \cmidrule(lr){5-8}
 & All & Legs & Vtx & Contact & Freq & Dist (mm) & Acc Err \\
\midrule
MVAE \cite{ling2020MVAE} & 26.1 & 32.3 & 44.6 & -    & 1.72\% & 1.4 &  6.74 \\
HuMoR \cite{rempe2021humor} & 22.7 & 26.1 & 35.5 & 0.97 & 1.18\% & 0.8 & 4.67 \\
PhaseMP \cite{shi2023phasemp} & 26.6 & 28.7 & 39.1 & 0.95 & 1.33\% & 1.2 & 6.47 \\
\rowcolor{gray!15} RoHM \cite{zhang2024rohm}  & 20.9 & 24.1 & 32.4 & 0.97 & \textbf{1.12}\% & 0.6 & 2.61 \\
MDM \cite{tevet2023human}  & 25.4 & 26.7 & 40.1 & 0.95 & 1.32\% & \textbf{0.5} & 6.24 \\
\midrule
VPoser-t \cite{pavlakos2019expressive} & 36.5 & 44.2 & 52.4 & -    & 1.53\% & 0.7 & 7.21 \\
PoseNDF-t \cite{tiwari22posendf}       & 24.6 & 27.5 & 32.7 & -    & 1.42\% & 0.9 & 6.97 \\
DPoser-t \cite{lu2023dposer}           & 25.1 & 27.9 & 32.4 & -    & 1.35\% & 0.8 & 7.13 \\
\rowcolor{gray!15} NRDF-t \cite{he24nrdf}                 & 22.8 & 22.4 & 28.1 & -    & 1.31\% & 0.7 & 6.89 \\
\midrule
DPoser + T-PoseNDF & 19.1 & 21.1 & 25.9 & 0.97 & 1.31\% & 0.7 & 3.29 \\
DPoser + T-NRDF    & 17.3 & 17.9 & 23.3 & 0.95 & 1.22\% & 0.5 & 3.13 \\
VPoser + T-PoseNDF & 21.4 & 23.5 & 31.6 & 0.97 & 1.20\% & 0.6 & 3.47 \\
VPoser + T-NRDF    & 19.0 & 20.8 & 27.5 & 0.96 & 1.29\% & 0.6 & 3.18 \\
PoseNDF + T-PoseNDF & 18.6 & 19.2 & 26.5 & 0.98 & 1.28\% & 0.6 & 3.24 \\
PoseNDF + T-NRDF   & 17.0 & 18.2 & 25.2 & 0.96 & 1.26\% & 0.7 & 3.08 \\
NRDF + T-PoseNDF   & 18.8 & 20.5 & 25.4 & 0.96 & 1.26\% & 0.7 & 3.19 \\
\rowcolor{gray!15} NRDF + T-NRDF      & 16.7 & 17.5 & 22.6 & 0.98 & 1.22\% & 0.5 & 2.97 \\
\midrule
{Motion-NDF}       & 18.1 & 19.8 & 23.5 & 0.98 & 1.22\% & 0.5 & 2.64 \\
\rowcolor{gray!15} \textbf{\name~(ours)} & \textbf{16.4} & \textbf{17.1} & \textbf{19.9} & \textbf{0.98} & 1.17\% & 0.5 & \textbf{2.25} \\
\bottomrule
\end{tabular}%
}\vspace{-6mm}
\end{table}
\paragraph{Baselines} %
As pose prior baselines, we employ VPoser \cite{pavlakos2019expressive}, DPoser \cite{lu2023dposer}, Pose-NDF \cite{tiwari22posendf}, and NRDF \cite{he24nrdf}. We also append \emph{-t} ({\eg} \emph{DPoser-t}) to indicate that only the corresponding $0$-order pose prior is used in the initialization phase, without the final optimization stage which involves higher-order priors ({\ie} transition \& acceleration prior). 
We also adopted the Euclidean PoseNDF \cite{tiwari22posendf} and geometric NRDF~\cite{he24nrdf} to model transitions (T-) and accelerations (A-). 
We compare our method with existing human motion priors \cite{tevet2023human,rempe2021humor,zhang2024rohm,shi2023phasemp,ling2020MVAE}. We further augment MDM \cite{tevet2023human} with infilling capabilities to tackle motion denoising. To achieve this, we train MDM with the same data split on AMASS. In ablations, we use the state-of-the-art human pose estimation (HPE) model SMPLer-X \cite{cai2023smplerx} as the initialization baseline and apply our motion priors to quantify the impact of each prior in optimization. 
For a fair comparison, all the methods we compared are trained and tested with the same data split as ours and \cite{tiwari22posendf,he24nrdf,pavlakos2019expressive}. For PROX dataset, we follow RoHM \cite{zhang2024rohm} to use off-the-shelf regressors \cite{li2022cliff,feng2021collaborative,sarandi2020metrabs} to obain per-frame initialization for a fair comparison. We provide more details in our supplementary material.

\paragraph{Evaluation metrics} To comprehensively evaluate our method, we categorize our metrics across multiple tasks:
\begin{itemize}
    \item \textbf{Motion estimation and refinement}: We measure global 3D joint positional error to evaluate the accuracy of estimated joint positions and meshes. We report the Mean Per Joint Position Error (MPJPE) and Procrustes-aligned Mean Per Joint Position Error (PA-MPJPE) in $mm$, and further distinguishing the MPJPE in the global coordinate system (GMPJPE) between visible (Vis), occluded (Occ), Leg (toes, ankles, and knees), all body parts (All), and vertices (Vtx) in $mm$.
    \item \textbf{Motion quality and plausibility}: We assess floor contact plausibility by reporting the frequency (Freq) and mean distance (Dist) of floor penetrations in $cm$, along with the overall contact rate (Contact). Mean per-joint acceleration $\|$Acc$\|$ and Acceleration Error (Acc Err) are measured in $mm/s^{2}$ to evaluate motion smoothness. To assess motion diversity and fidelity, we report Average Pairwise Distance (APD) and Fréchet Distance (FID), respectively. Specifically, we evaluate FID for pose (FID$_{p}$) and for motion (FID$_{m}$) following \cite{he2022nemf}.
\end{itemize}

\subsection{Evaluation \& Results}\label{sec:opt_eval}
As introduced in \cref{sec:NRMF}, with accurate learned neural (Riemannian) distance fields, we are able to map any noisy motion observations ({\eg} noisy 3D mocap data) onto a plausible one through optimization with our motion prior while maintaining the alignment with the original observation. In this section, we leverage {\name} as a set of motion prior terms in an optimization-based pipeline for various downstream tasks from different observations, such as motion estimation from RGB(-D) based 2D\&3D observations, and motion refinement. 

\paragraph{Motion estimation from 3D observations} 
For a fair comparison, we follow \cite{rempe2021humor} to extract (i) noisy and (ii) partial 3D keypoints from the AMASS dataset and sample $3s$ (90 frames). We conduct both the motion estimation accuracy and plausibility evaluation of (i) motion denoising and (ii) fitting to partial 3D observation.

\begin{table}[t]
\centering
\caption{{Motion and shape estimation from partial 3D observations.}}
\label{tab:partial}
\resizebox{\linewidth}{!}{%
\footnotesize
\begin{tabular}{lccccccccc}
\toprule
\textbf{Method} & \multicolumn{5}{c}{\textbf{Positional Error (mm)}} & \multicolumn{4}{c}{\textbf{Motion Prop.}} \\ 
\cmidrule(lr){2-6} \cmidrule(lr){7-10}
 & Vis & Occ & All & Legs & Vtx & Contact & Freq & Dist (mm) & Acc Err \\ 
\midrule
MVAE \cite{ling2020MVAE} & 23.9 & 191.5 & 105.6 & 188.6 & 114.5 & - & 3.15\% & 3.0 & 6.91 \\ 
HuMoR (VPoser) \cite{rempe2021humor} & 14.6 & 120.3 & 67.9 & 118.1 & 76.8 & 0.89 & 3.31\% & 2.6 & 4.78 \\ 
PhaseMP \cite{shi2023phasemp} & 38.7 & 177.5 & 110.1 & 175.5 & 118.4 & 0.89 & 3.28\% & 3.1 & 6.29 \\
\rowcolor{gray!15} RoHM \cite{zhang2024rohm}  & 12.9 & 88.4 & 51.3 & 86.7 & 58.4 & 0.91 & 3.08\% & \textbf{1.6} & 2.43 \\
MDM \cite{tevet2023human}  & 27.8 & 173.6 & 101.2 & 170.6 & 112.7 & 0.88 & 3.18\% & 2.6 & 6.23 \\
\midrule
VPoser-t \cite{pavlakos2019expressive} & 6.7 & 207.6 & 108.8 & 205.4 & 118.3 & - & 16.77\% & 22.8 & 7.18 \\ 
PoseNDF-t \cite{tiwari22posendf} & 5.9 & 204.1 & 104.6 & 200.9 & 113.7 & - & 15.49\% & 23.1 & 7.01 \\ 
DPoser-t \cite{lu2023dposer} & 4.9 & 211.2 & 109.4 & 209.3 & 117.5 & - & 15.65\% & 22.5 & 6.95 \\ 
\rowcolor{gray!15} NRDF-t \cite{he24nrdf} & \textbf{4.2} & 194.5 & 100.2 & 192.7 & 109.1 & - & 15.23\% & 20.9 & 6.73 \\ 
\midrule
DPoser + T-PoseNDF & 11.2 & 111.5 & 61.8 & 109.2 & 72.4 & 0.86 & 3.67\% & 3.4 & 3.36 \\
DPoser + T-NRDF & 11.0 & 102.8 & 56.7 & 100.8 & 66.3 & 0.87 & 3.12\% & 2.9 & 3.19 \\
VPoser + T-PoseNDF & 13.4 & 108.7 & 60.5 & 106.9 & 71.7 & 0.88 & 3.29\% & 3.1 & 3.41 \\ 
VPoser + T-NRDF & 12.0 & 102.3 & 57.8 & 100.2 & 67.9 & 0.87 & 3.19\% & 2.2 & 3.13 \\
PoseNDF + T-PoseNDF & 11.5 & 100.2 & 56.3 & 97.6 & 65.8 & 0.88 & 3.17\% & 2.8 & 3.35 \\ 
PoseNDF + T-NRDF & 11.6 & 97.3 & 54.9 & 95.4 & 64.2 & 0.88 & 3.23\% & 2.6 & 3.11 \\
NRDF + T-PoseNDF & 11.1 & 108.1 & 59.1 & 105.8 & 69.5 & 0.87 & 3.16\% & 2.5 & 3.27 \\
\rowcolor{gray!15} NRDF + T-NRDF & 10.5 & 87.1 & 48.5 & 84.8 & 58.6 & 0.89 & 3.17\% & 2.1 & 3.01 \\
\midrule
Motion-NDF & 11.6 & 99.5 & 55.2 & 97.3 & 63.9 & 0.89 & 3.17\% & 2.6 & 2.72 \\
\rowcolor{gray!15} \textbf{\name~(ours)} & 9.7 & \textbf{83.8} & \textbf{46.3} & \textbf{81.6} & \textbf{56.7} & \textbf{0.89} & \textbf{3.05\%} & 1.9 & \textbf{2.31} \\ 
\bottomrule
\end{tabular}%
}\vspace{-3mm}
\end{table}

To perform motion denoising, we align the recovered human body well with the observations and preserve the plausibility of the motion. To test this, we sample \textcolor{steelblue}{\textbf{noisy keypoints}} from the AMASS dataset with a Gaussian noise of $4cm$. As shown in \cref{tab:denoising}, we compare {\name} with state-of-the-art pose \cite{tiwari22posendf, he24nrdf, lu2023dposer, pavlakos2019expressive} and motion priors \cite{rempe2021humor,ling2020MVAE,zhang2024rohm,shi2023phasemp,tevet2023human}. It can be seen that our method consistently outperforms the {\sota}. Previous methods typically only focus on $0$-order or $1^{\text{st}}$-order cues either neglecting the high-order information \cite{pavlakos2019expressive,tiwari22posendf,he24nrdf,lu2023dposer} thus producing unrealistic, oversmooth, and implausible motion transitions, or causing error accumulation over time (drift) \cite{rempe2021humor}. In contrast, our method incorporates priors for all essential components of dynamics to fill in the plausibility gap while maintaining the fidelity of each frame, which is supported by the comparison of the output distribution in \cref{fig:all_distribs}. Moreover, as shown in \cref{tab:denoising}, methods that properly treat the geometry of articulations tend to perform better, while PoseNDF based (Euclidean) projections fail under larger noise. %
\setlength{\columnsep}{6pt}%
\begin{wrapfigure}[13]{R}{0.5\columnwidth}
            \vspace{-0.8cm}
            
            \begin{center}
    \includegraphics[width=\linewidth]{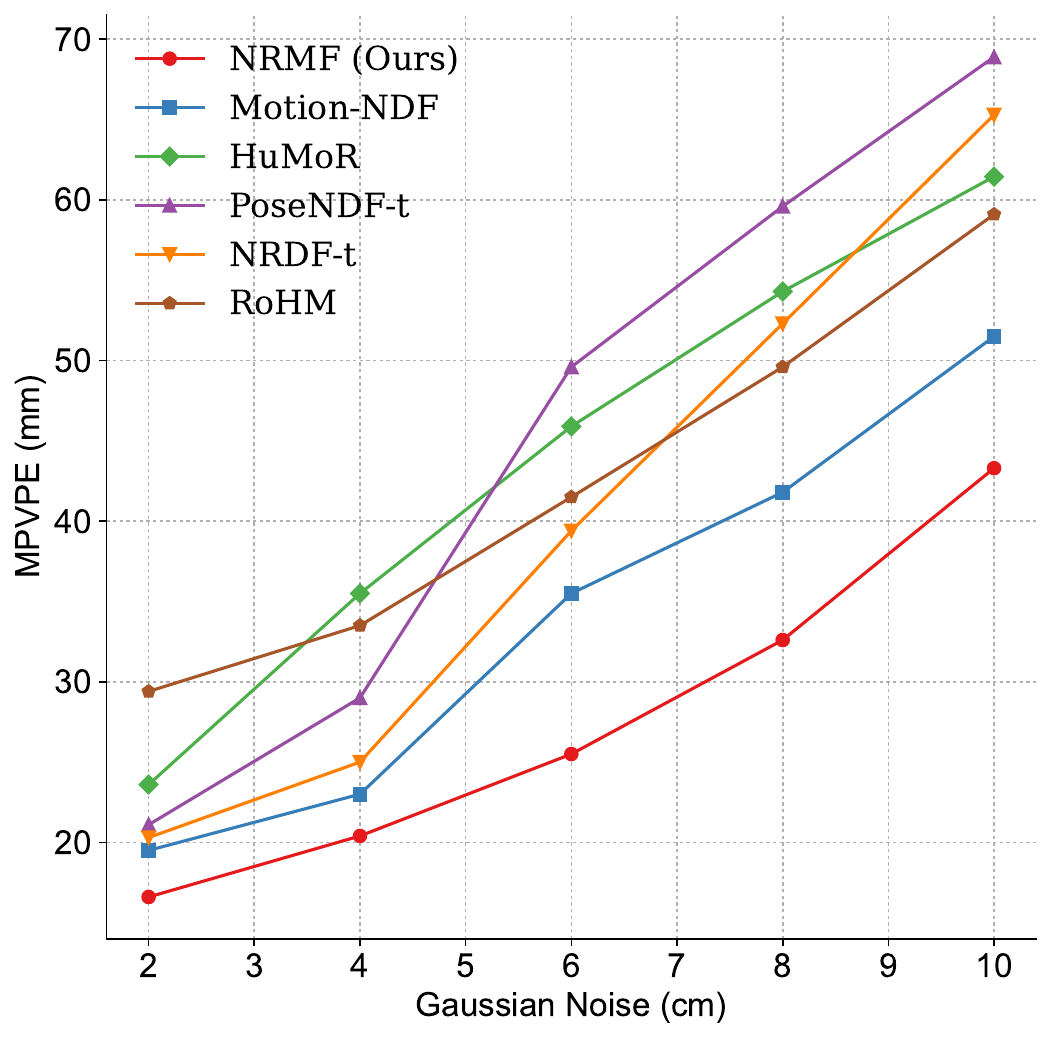}
            \end{center}
            \vspace{-6mm}
            \caption{Behavior of different methods for increasing levels of noise.}
    \label{fig:noise_curves}
\end{wrapfigure} 
MDM \cite{tevet2023human} and RoHM \cite{zhang2024rohm} present comparable results for contact plausibility but yield low estimation accuracy. \cref{fig:noise_curves} demonstrates the effect of gradual (Gaussian) noising on MPVPE, where our method consistently performs better and behaves more stable under increased noise.

\begin{figure}[t]
        \centering
        \includegraphics[width=\linewidth]{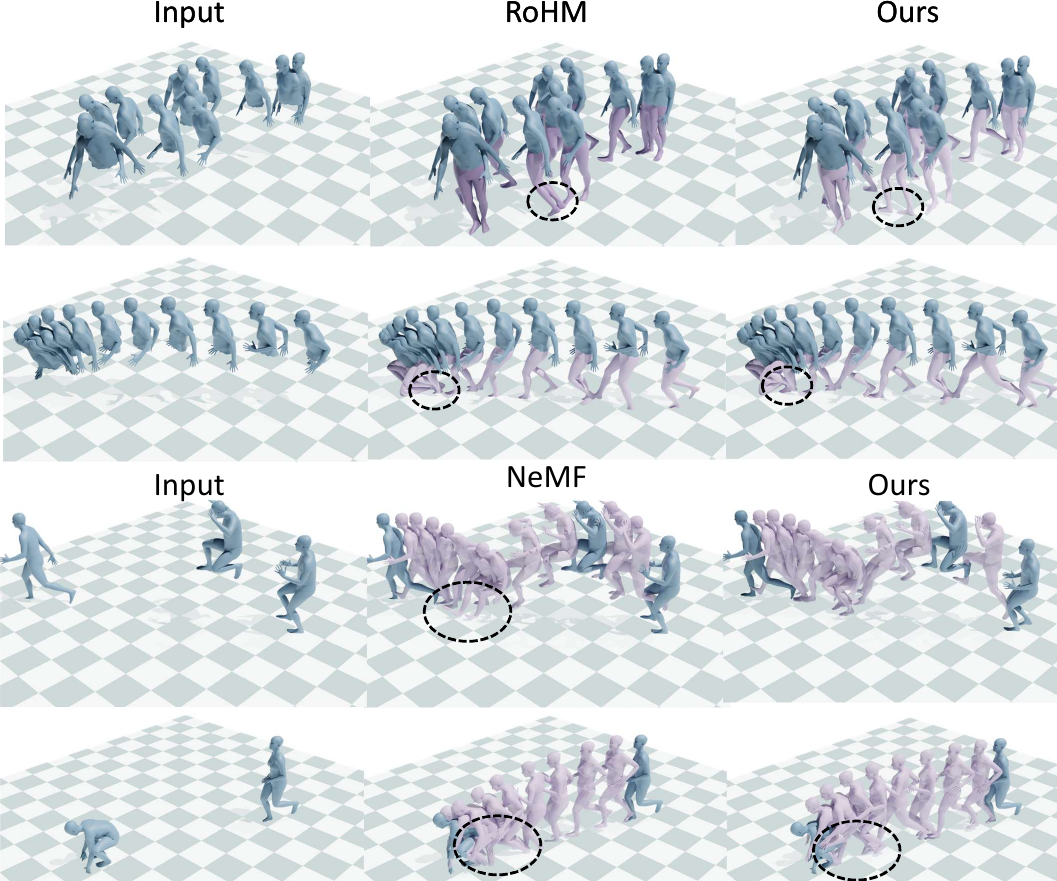}
        \caption{ {Qualitative comparison with state-of-the-art methods} \cite{zhang2024rohm,he2022nemf} on motion estimation from \textcolor{steelblue}{\textbf{noisy partial}} observation (top two) and motion in-between (bottom two) where the \textcolor{SoftLavender}{\textbf{purple body parts}} represent the recovered motion. \vspace{-4mm}}
        \label{fig:qua} 
\end{figure}

To test motion estimation under occlusions (\ie \textcolor{steelblue}{\textbf{partial observation}})  we sample the AMASS dataset with a standard mask height of $0.9m$ following \cite{rempe2021humor} to yield partial 3D observations.
\cref{tab:partial} reveals that while applying only a pose prior term ({\ie} \emph{-t}) 
gives lower errors for visible body parts, the model suffers from over-smoothing and higher reconstruction and acceleration errors when deducing invisible body parts over time, due to the inherent negligence of transitions. 
In contrast, our {\name}~consistently and significantly outperforms both baselines and prior state-of-the-art methods. This demonstrates the effectiveness of our approach in modeling motion dynamics: by jointly incorporating pose, transition, and acceleration priors, {\name}~achieves improved pose accuracy while producing more temporally coherent and physically plausible motions, and successfully mitigates error accumulation. Qualitative results in \cref{fig:qua} and Supp. Mat. further support these observations.

\paragraph{Motion refinement for human mesh recovery} 
\begin{table}[t]
\centering
\caption{{Ftting to 2D obsertations on 3DPW.} We compare the refinement results after optimizing SMPLer-X \cite{cai2023smplerx} with different prior terms. MPJPE and MPVPE are in millimeters.\vspace{-2mm}}\label{tab:3DPW}
\resizebox{\linewidth}{!}{
\begin{tabular}{l c c c c c}
\toprule
\textbf{Method} & \textbf{MPJPE (mm)} & \textbf{MPVPE (mm)} & \textbf{Acc Err ($m/s^{2}$)} & \textbf{Trans Err ($\times 10^{-3}$)} \\
\midrule
SMPLer-X \cite{cai2023smplerx} & 82.65 & 94.23 & 23.71 & 31.63 \\
+ No prior & 84.67 & 96.82 & 26.75 & 34.54 \\
+ VPoser \cite{pavlakos2019expressive} & 79.98 & 91.53 & 25.82 & 30.58  \\
+ DPoser \cite{lu2023dposer} & 75.45 & 87.04 & 27.54 & 28.69 \\
+ PoseNDF \cite{tiwari22posendf} & 73.49 & 84.61 & 25.12 & 29.77 \\
+ NRDF \cite{he24nrdf} & 71.88 & 83.23 & 24.31 & 26.38 \\
+ T-NRDF & 66.98 & 76.94 & 9.89 & 7.98 \\
+ A-NRDF & 70.88 & 81.92 & 6.73 & 11.87 \\
\midrule
+ RoHM \cite{zhang2024rohm} & 69.78 & 79.72 & 9.13 & 12.37 \\
\rowcolor{gray!15} + \textbf{\name} (full) & \textbf{66.13} & \textbf{75.61} & \textbf{6.52} & \textbf{5.67} \\
\bottomrule
\end{tabular}}
\vspace{-3mm}
\end{table}
Our motion prior can also be used with regression-based human pose estimation methods as an efficient optimization-based refinement module for mesh recovery from videos. \cref{tab:3DPW} compares our method with the {\sota} baselines of \cite{zhang2024rohm,he24nrdf,pavlakos2019expressive,lu2023dposer} using SMPLer-X \cite{cai2023smplerx} and \cite{cao2019openpose} as initialization.
Notably, while optimization improves pixel alignment, it does not guarantee plausible motion due to the lack of constraints and biased optimization without appropriate priors.
Specifically, only applying a pose prior ($0$-order) usually leads to large errors in transition and acceleration due to the absence of higher-order motion constraints. Notably, the ablations highlight the significance of each prior term separately. 
By integrating acceleration and transition, we improve the Acc Err and Trans Err significantly. Applying our \textbf{\name (full)} model for refinement will boost the per-frame mesh reconstruction accuracy while preserving plausible and realistic motion transitions. %

\begin{figure}[t]
        \centering
        \includegraphics[width=\linewidth]{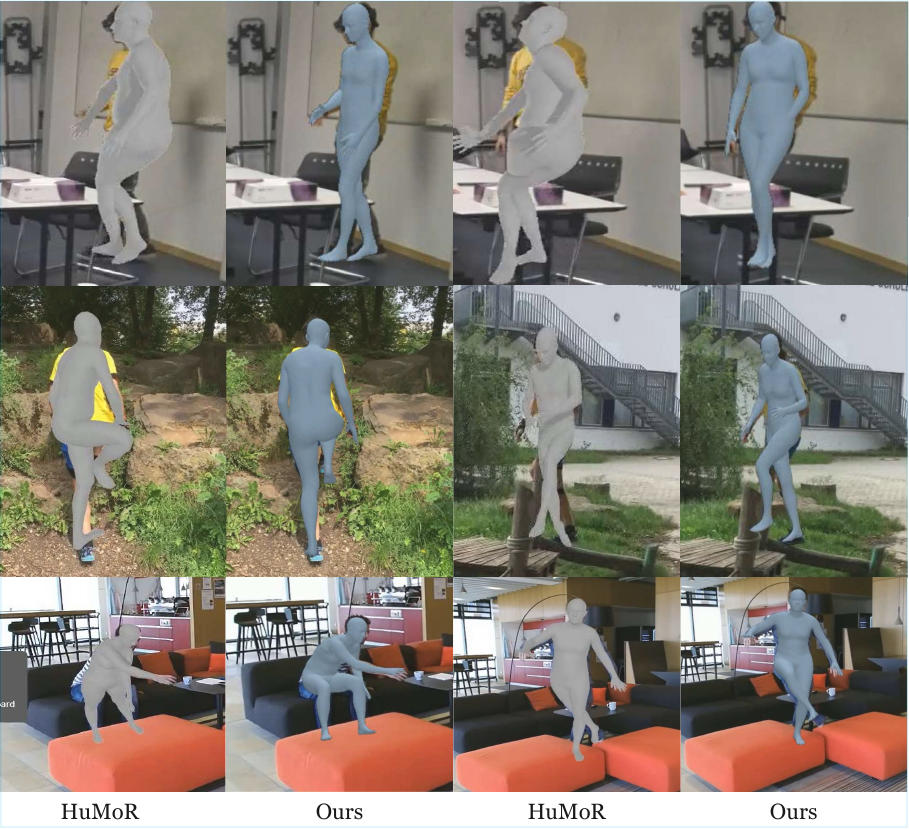}
        \caption{ {Qualitative comparison with state-of-the-art methods} for in-the-wild motion estimation.}
        \label{fig:qua_small} 
\end{figure}

\paragraph{Motion estimation from RGB(-D) in-the-wild observations}
As discussed in \cref{sec:NRMF}, our method is readily applicable to real-world cases, enabling robust motion recovery from various modalities including in-the-wild RGB(-D) videos, \ie partial and noisy 2D \& 3D observations. We conduct our evaluation on i3DB (RGB), EgoBody (RGB) and PROX (RGB \& RGB-D) datasets. Specifically, we leverage the off-the-shelf 2D pose estimation \cite{cao2019openpose}, human mask segmentation \cite{chen2017rethinking}, along with a plane detection \cite{liu2019planercnn} model as observations during optimization following \cite{rempe2021humor}. 
As shown in \cref{tab:i3db,tab:prox,tab:egobody}, conventional pose priors perform reasonably well on visible joints but struggle in occluded regions. Relying solely on $0$-order pose information leads them to overfit visible joints, producing temporally inconsistent and physically implausible reconstructions. In contrast, {\name} incorporates richer dynamics through transition and acceleration priors, yielding substantially lower position and acceleration errors on both visible and occluded joints compared to existing motion~\cite{rempe2021humor,zhang2024rohm} and pose~\cite{he24nrdf,tiwari22posendf,lu2023dposer} priors. Qualitative comparisons in \cref{fig:qua_small} highlight our method’s superior motion recovery, while \cref{fig:all_distribs} shows that the pose and transition distributions learned by {\name} better match the ground-truth distribution than those of prior models. We provide more details in our supplementary material as well as more in-the-wild qualitative results in the \textcolor{magenta}{supplementary videos}.

\paragraph{Motion Generation and In-betweening}%
\begin{table}[t]
    \centering
    \caption{{Quantitative analysis of pose and motion generation.}\vspace{-2mm}}\label{tab:gen}
    \resizebox{\linewidth}{!}{
    \footnotesize
    \setlength{\tabcolsep}{10pt}
    \vspace{-2mm}
    \begin{tabular}{lcccc}
        \toprule
        \textbf{Method} & \textbf{APD $\uparrow$} (cm) & \textbf{FID$_{p}$ $\downarrow$} & \textbf{FID$_{m}$ $\downarrow$} & $\|$\textbf{Acc}$\|$ $\downarrow$ \\
        \midrule
        GMM \cite{bogo2016keep} & 16.28 & 0.435 & -  & - \\
        GAN-S \cite{davydov2022adversarial} & 15.68 & 0.201 & -  & - \\
        VPoser \cite{pavlakos2019expressive} & 10.75 & \textbf{0.113} & -  & - \\
        Pose-NDF \cite{tiwari22posendf} & 18.54 & 1.124 & -  & - \\
        DPoser \cite{lu2023dposer}  & \textbf{36.25} & 2.344 & -  & - \\
        NRDF \cite{he24nrdf} & 24.87 & 0.706 &  -  & - \\
        \midrule
        MVAE \cite{ling2020MVAE} & 85.26 & 1.475 & 10.571  & 8.15 \\
        HuMoR \cite{rempe2021humor}  & \textbf{98.47} & \textbf{0.278} & 9.147  & 4.96 \\
        Motion-NDF & 93.46 & 0.894 & 8.351  & - \\
        \rowcolor{gray!15} \textbf{\name}~(ours) & 96.37 & 0.718 & \textbf{5.317}  & \textbf{3.18} \\
        \bottomrule
    \end{tabular}}
    \vspace{-3mm}
\end{table}
\begin{table}[t]
\centering
\caption{{Motion and shape from RGB sequences on i3DB.}}\label{tab:i3db}
\vspace{-2mm}
\resizebox{\columnwidth}{!}{  
\begin{tabular}{lcccccccc}
\toprule
\textbf{Method} & \multicolumn{4}{c}{\textbf{Global Joint Error (cm)}} & \multicolumn{4}{c}{\textbf{Root-Aligned Joint Error (cm)}} \\
\cmidrule(lr){2-5} \cmidrule(lr){6-9}
 & Vis & Occ & All & Legs & Vis & Occ & All & Legs \\
\midrule
VIBE \cite{kocabas2019vibe} & 90.05 & 192.55 & 116.46 & 121.61 & 12.06 & 23.78 & 15.08 & 21.65 \\
MVAE \cite{ling2020MVAE} & 37.54 & 50.63 & 40.91 & 44.42 & 16.00 & 28.32 & 19.17 & 26.63 \\
\rowcolor{gray!15} HuMoR \cite{rempe2021humor} & 26.37 & 34.79 & 28.22 & 31.41 & 12.02 & 21.73 & 14.51 & 20.84 \\
\midrule
VPoser-t & 28.33 & 40.97 & 31.59 & 35.16 & 12.77 & 26.48 & 16.31 & 25.64 \\
PoseNDF-t & 29.17 & 42.36 & 32.64 & 36.28 & 13.19 & 27.12 & 17.03 & 26.14 \\
\rowcolor{gray!15} NRDF-t & 27.85 & 39.48 & 30.82 & 34.13 & 12.53 & 25.88 & 15.92 & 24.89 \\
DPoser-t & 28.12 & 40.23 & 31.32 & 34.74 & 12.63 & 26.05 & 16.17 & 25.08 \\
\midrule
\rowcolor{gray!15} \textbf{\name} (ours) & \textbf{22.35} & \textbf{25.12} & \textbf{24.67} & \textbf{23.68} & \textbf{10.27} & \textbf{17.51} & \textbf{11.97} & \textbf{16.73} \\
\bottomrule
\end{tabular}
}\vspace{-2mm}
\end{table}
For motion generation, we follow the protocol of~\cite{rempe2021humor}, sampling 50 motion sequences ($2\mathrm{sec}$ each) for each initial state. As shown in \cref{tab:gen}, VPoser achieves the lowest FID and APD scores, but produces less diverse and expressive motions. In contrast, DPoser yields high diversity (APD) at the cost of low plausibility (FID). NRDF strikes a balance between fidelity and diversity. Among the motion priors, our method achieves the best overall FID for motion while maintaining competitive APD, indicating that {\name} generates realistic and diverse motions. Additionally, {\name} produces the lowest mean acceleration norm, reflecting stable motion with reduced jitter.
For in-betweening, we follow a similar protocol to \cite{he2022nemf} and \cite{Dang2024MoManifold}, focusing on sparse keyframe supervision and infilling refinement. Specifically, we provide a sparse set of keyframe observations corresponding either to the two fixed endpoints, to $10\%$ and $20\%$ of the total frames, or to a $30\%$ scheme divided across the start, middle, and end of the sequence similar to the protocol in \cite{tevet2023human}. We compare our method against \cite{rempe2021humor,he2022nemf} and SLERP. Our method explicitly leverages high-order motion dynamics, enabling robust interpolation even under extremely sparse constraints. \cref{fig:qua} illustrates a qualitative comparison in the two-endpoint setting, where our method produces more natural and plausible interpolations. Additional qualitative results, as well as quantitative evaluations across different sparsity levels, are provided in the supplementary materials.

\begin{table}[t]
\centering
\caption{{Motion estimation from RGB(-D) input on PROX. }}
\vspace{-2mm}
\resizebox{\columnwidth}{!}{  
\begin{tabular}{lcccccccc}
\toprule
\textbf{Method} & \multicolumn{4}{c}{\textbf{RGB-D}} & \multicolumn{4}{c}{\textbf{RGB}} \\
\cmidrule(lr){2-5} \cmidrule(lr){6-9}
 & $\|$Acc$\|$ & Dist & FID$_{m}$ & Freq & $\|$Acc$\|$ & Dist & FID$_{m}$ & Freq \\
\midrule
HuMoR \cite{rempe2021humor} & 1.88 & 35.67 & 11.65 & 6.13\% & 2.35 & 41.92 & 12.47 & 10.52\% \\
PhaseMP \cite{shi2023phasemp} & - & - & - & - & 1.80 & 46.96 & - & -\\
RoHM \cite{zhang2024rohm} & 1.79 & 6.67 & 12.16 & 3.21\% & 2.19 & \textbf{14.31} & 13.65 & 4.65\% \\
\midrule
VPoser-t & 3.45 & 53.49 & 21.89 & 10.66\% & 3.24 & 54.96 & 19.64 & 13.95\% \\
PoseNDF-t & 2.84 & 48.77 & 16.45 & 11.35\% & 2.93 & 50.13 & 16.87 & 13.46\% \\
NRDF-t & 2.67 & 45.91 & 14.58 & 9.97\% & 2.77 & 46.78 & 14.32 & 12.98\% \\
DPoser-t & 3.32 & 49.45 & 17.32 & 10.89\% & 3.08 & 51.35 & 16.98 & 12.21\% \\
\midrule
\rowcolor{gray!15} \textbf{\name} (Ours) & \textbf{1.48} & \textbf{5.78} & \textbf{8.95} & 2.78\% & \textbf{1.73} & 16.21 & \textbf{14.13} & 5.12\% \\
\bottomrule
\end{tabular}
}\vspace{-2mm}
\label{tab:prox}
\end{table}

\begin{table}[t]
\centering
\caption{\label{tab:egobody}{Motion estimation results on EgoBody dataset}.\vspace{-3mm}}
\resizebox{\linewidth}{!}{%
\begin{tabular}{lccccccccc}
\toprule
\textbf{Method} & \multicolumn{4}{c}{\textbf{Positional Error (cm)}} & \multicolumn{4}{c}{\textbf{Motion Prop.}} \\ 
\cmidrule(lr){2-5} \cmidrule(lr){6-9}
 & Vis & Occ & Legs & All & Vtx & Freq & Dist & Acc \\ 
\midrule
HuMoR \cite{rempe2021humor} & 75.3 & 129.5 & 124.6 & 108.85 & 105.94 & 7.31\% & 12.65 & 5.36 \\ 
RoHM \cite{zhang2024rohm}  & 60.1 & 123.6 & 116.9 & 98.33 & 93.81 & 6.08\% & \textbf{2.15} & 2.97 \\
\midrule
VPoser-t \cite{pavlakos2019expressive} & 65.5 & 139.4 & 129.7 & 110.30 & 116.14 & 21.77\% & 17.34 & 7.18 \\ 
PoseNDF-t \cite{tiwari22posendf} & 62.1 & 135.6 & 131.4 & 106.41 & 99.05 & 19.49\% & 16.98 & 7.01 \\ 
DPoser-t \cite{lu2023dposer} & 68.1 & 141.3 & 138.6 & 112.14 & 118.47 & 22.65\% & 18.35 & 6.95 \\ 
\rowcolor{gray!15} NRDF-t \cite{he24nrdf} & 60.3 & 131.6 & 133.6 & 103.03 & 109.22 & 19.23\% & 15.26 & 6.73 \\ 
\midrule
\rowcolor{gray!15} \textbf{\name} (Ours) & \textbf{57.5} & \textbf{112.6} & \textbf{109.8} & \textbf{92.10} & \textbf{90.06} & \textbf{4.78\%} & 3.78 & \textbf{1.58} \\ 
\bottomrule
\end{tabular}%
}\vspace{-3mm}
\end{table}

%% file: sec/7_conclusion.tex
\section{Conclusion}
In summary, our work presented Neural Riemannian Motion Fields (\name), a novel framework for modeling 3D human motion while rigorously respecting the underlying geometry of articulated motions. By decomposing motion into its constituent components—pose, transition, and acceleration—and representing these as the zero-level sets of dedicated neural distance fields, \name~acts as a strong inductive bias bridging the gap between learning and physical modeling of human motion. Our proposed projection and geometric integration algorithms help ensuring temporal consistency and robustness against noise and enable versatile applications ranging from motion denoising and in-betweening to test-time optimization for mesh recovery. Extensive experiments on diverse benchmarks consistently confirm that \name~outperforms existing methods, delivering enhanced accuracy and more lifelike trajectories. 

\paragraph{Limitation \& future work}
Our iterative updates can incur significant runtime leading to minutes for processing a sequence of $<1\mathrm{min}$. Moreover, our projected-integrators lack rigorous theoretical understanding, preventing us from certifying their optimality. As such, our work leaves ample room for future study. \emph{Learning-to-optimize} can be an effective future direction, while principled sampling algorithms such as \emph{Riemannian Langevin MCMC}~\cite{birdal2019probabilistic} can be employed as integrators. 

\paragraph{Acknowledgement} 
T. Birdal acknowledges support from the Engineering and Physical Sciences Research Council [grant EP/X011364/1]. 
T. Birdal was supported by a UKRI Future Leaders Fellowship [grant MR/Y018818/1].

%% file: sec/X_suppl.tex
\clearpage
\setcounter{page}{1}
\setcounter{prop}{0}
\setcounter{section}{0}
\renewcommand\thesection{\Alph{section}}
\maketitlesupplementary

\section{Geometry of Articulated Motions}
For the sake of a self-contained manuscript, we now provide a complete treatment of the Riemannian geometry of rotations that we use in this work. For completeness, we also recall the definitions we provide in the main paper. 

\paragraph{Riemannian manifolds}
Following~\cite{birdal2018bayesian,birdal2019probabilistic,chen2022projective}, we define an $m$-dimensional \textit{Riemannian manifold}, embedded in an ambient Euclidean space $\Amb = \R^d$ and endowed with a \textit{Riemannian metric} $\G\triangleq (\Gx)_{\x\in\Man}$ to be a smooth curved space $(\Man,\G)$. A vector $\v\in\Amb$ is said to be \emph{tangent} to $\Man$ at $\x$ \emph{iff} there exists a smooth curve $\curve:[0,1]\to\Man$ s.t. $\curve(0)=\x$ and $\dcurve(0)=\v$. The velocities of all such curves through $\x$ form the \emph{tangent space} $\TxM=\{ \dcurve (0) \,|\, \curve:\R\to\Man \text{ is smooth around $0$ and } \curve(0)=\x\}$.
We will also work with a product of $K$ manifolds, $\Man_{1:K}\vcentcolon=\Man_1\times\Man_2\times\dots\times\Man_K$, For identical manifolds, \ie $\Man_i\equiv\Man_j$, we recover the \emph{power manifold}, $\Man^K\vcentcolon=\Man_{1:K}$. In this work, we will specifically work on the differentiable manifold of a product of $K$ rotations, $\SO^K$.

\begin{dfn}[$\SO$] The manifold of 3D rotations admit the structure:
\begin{equation}
\SO=\left\{\Rot \in \R^{3 \times 3}: \Rot^{\top} \Rot=\mathbf{I}, \operatorname{det}(\Rot)=1\right\}.
\end{equation}    
\end{dfn}
To characterize this manifold, we start by deriving its tangent space:
\begin{dfn}[Tangent space of $\SO$]
The tangent space of $\SO$ at $\Rot$ is defined throughout as the \emph{left-invariant vector fields}:
\begin{equation}
    \TSOR=\left\{\Rot\Skew \,:\, \Skew\in\solie\right\}, 
\end{equation}
where $\solie$ is given by the set of \emph{skew-symmetric} tensors:
\begin{equation}
    \solie=\left\{\Skew \in \R^{3 \times 3}: \Skew^{\top}=-\Skew\right\}, 
\end{equation}
and $\Skew\hb = \rod \times \hb$ for any $\hb\in\R^3$. $\rod$ is called the \emph{axial} or \emph{rotation vector}. 
\end{dfn}
\begin{proof}
Differentiating the constraints that specify $\SO$ leads us to the definition of skew-symmetric matrices:
\begin{align}
        \frac{\diff(\Rot^\top\Rot)}{\diff t} = \dRot^\top\Rot + \Rot^\top \dRot  = 0 \implies \Skew = -\Skew^\top, %
    \end{align}
    where $\Skew=\Rot^\top\dRot$ is a skew symmetric matrix and the vector $\dRot=\Rot\Skew$ is tangent to $\Rot$. 
\end{proof}
To work with functions on this manifold, for example in optimization, we will further introduce the notions of differentials and gradients that makes intrinsic sense.
\begin{dfn}[Riemannian gradient for $\SO$]
The Riemannian (or intrinsic) gradient of a function defined on $\SO$ is the unique tangent vector satisfying:
    \begin{equation}
    \langle\grad f(\Rot), \xi\rangle_R=D f(\Rot)[\bxi]=\langle\nabla f(\Rot), \bxi\rangle \quad
    \end{equation}
    for all $\bxi \in \TR \SO$. 
\end{dfn}
The intrinsic gradient is the only part of the gradient that makes sense to a local citizen. For $\SO$, given the Euclidean gradient, it can be obtained via the $\mathrm{egrad2rgrad}$ operator:
\begin{prop}[$\mathrm{egrad2rgrad}$ for $\SO$]
The projection of the Euclidean gradient $\nabla f(\Rot)$ onto the tangent space of $\Rot$ yields the Riemannian gradient:
\begin{align}
    \grad f(\Rot)&\triangleq \Pi_{\Rot}\big( \nabla f(\Rot)\big)\\
    &=\Rot\,\skewrm\left(\Rot^\top \nabla f\left(\Rot\right)\right)\\
    &=\frac{1}{2} \Rot\left(\Rot^\top \nabla f(\Rot)-\nabla f(\Rot)^\top \Rot\right),
\end{align}
where $\Pi_{\Rot}:\Amb\to\TSOR$ is an orthogonal projector onto the tangent space of $\Rot$.
This is famously known as the $\mathrm{egrad2rgrad}$ in various packages including ManOpt~\cite{pymanopt}.
\end{prop}
\begin{proof}
In a differential geometric setting, any vector can be written down as a linear combination of normal and tangential components.
To project the Euclidean gradient onto the tangent space
$\TSOR=\left\{\Rot\Skew \,:\, \Skew\in\solie\right\}$, our aim will be to \emph{kill} the normal component, leaving only the tangential one.
Let us start by writing:
\begin{equation}
 \nabla f\left(\Rot\right)=\Rot\left(\Rot^\top\nabla f\left(\Rot\right)\right).   
\end{equation}
Since any square matrix can uniquely be written as the sum of a symmetric and a skew-symmetric matrix, we can write:
\begin{align}
    \nabla f\left(\Rot\right) &= \Rot\sym\left(\Rot^\top\nabla f\left(\Rot\right)\right)+\Rot\skewrm\left(\Rot^\top\nabla f\left(\Rot\right)\right).\nonumber
\end{align}
By definition of the tangent space, $\Rot\skewrm(\vb)\in\TSOR$ for any $\vb\in\R^3$. Also note that $\Rot\sym\left(\Rot^\top\nabla f\left(\Rot\right)\right)$ lives in the \emph{normal space}, \ie, it is orthogonal to all tangent vectors with respect to the Frobenius inner product. To see this, let $\Sb=\sym\left(\Rot^\top\nabla f\left(\Rot\right)\right)$ and write:
\begin{align}
\langle \Rot\Sb,\Rot\Skew \rangle &= \trace\left(\left(\Rot\Sb\right)^\top\left(\Rot\Skew\right)\right)\\
&= \trace\left(\Sb^\top\Rot^\top\Rot\Skew\right)\\
&= \trace\left(\Sb^\top\Skew\right)\\
&= 0.
\end{align}
Note once again that $\Rot\Skew\in\TSOR$ by definition. 
To see the last equality, consider
\begin{align}
    \trace(\Sb \Skew)=\trace\left((\Sb \Skew)^\top\right)=\trace((-\Skew) \Sb)=-\trace(\Sb \Skew).\nonumber
\end{align}
Observe that $\trace(\Sb \Skew)=-\trace(\Sb \Skew) \implies \trace(\Sb \Skew)=0$.
Therefore, because $\Rot\sym\left(\Rot^\top\nabla f\left(\Rot\right)\right)$ is normal, it vanishes under projection onto the tangent space and we are left with the skew component, $\Rot\skewrm\left(\Rot^\top\nabla f\left(\Rot\right)\right)$, leading to the following orthogonal projection:
\begin{align}
    \grad f(\Rot)&\triangleq \Pi_{\Rot}\big( \nabla f(\Rot)\big)\\
    &= \Rot\skewrm\left(\Rot^\top\nabla f\left(\Rot\right)\right).
    \end{align}
This concludes the proof.
\end{proof}
Finally, we will introduce the exponential and logarithmic maps that are required to \emph{walk} on $\SO$. Most of the proofs, which we will omit for brevity, are based on the properties of the \emph{matrix exponential}. 
\begin{dfn}[Exponential Map ($(\Exp)$)]
    Let $\bxi\in\TSOR=\Rot\Skew$ denote any tangent vector with $\Skew=[\rod]_{\times}$. The exponential map at $\Rot$ is defined as a map $\Exp:\SO\times \TSOR\to \SO$:
    \begin{align}
        \Exp_{\Rot}(\Skew) &= \Rot\Exp(\Skew)\\
        &=\Rot\left(\Id+\frac{\sin(\theta)}{\theta}\Skew+\frac{1-\cos(\theta)}{\theta}\Skew^2\right),
    \end{align}
    where $\theta=\|\omega\|$ and $\Exp(\Skew)$ (right hand side) is known as the celebrated \textbf{Rodrigues’ formula}.
\end{dfn}

\begin{dfn}[Logarithmic Map ($(\Log)$)]
    The inverse of $\Exp$ is given by the logarithmic map $\Log:\SO\times\SO\to\TSOR$, which expresses any rotation $\Rot_1$ near $\Rot$ in the tangent space $\TSOR$:
    \begin{align}
    \Skew&=\Log_\Rot(\Rot_1) = \Log(\Rot^\top\Rot_1)\\
    &=
    \begin{cases}
        \frac{\theta}{2\sin(\theta)}\left( \Rot^\top\Rot_1 - \Rot_1^\top\Rot \right), &\theta\in(\pi, \pi) \setminus \{0\}\\
        \frac1{2}\left( \Rot^\top\Rot_1 - \Rot_1^\top\Rot \right), &\theta = 0,
    \end{cases}\nonumber
    \end{align}
    where
    \begin{align}
        \theta := d_{\SO}(\Rot,\Rot_1) = \cos^{-1}\left(\frac{\trace\left(\Rot^\top\Rot_1\right)-1}{2}\right),
    \end{align}
    is known as the \textbf{geodesic distance} between two rotations.
\end{dfn}

Our approach benefits from the dynamics on $\SO$ through the angular velocity and acceleration, which are the backbone of geometric mechanics~\cite{holm2011geometric,holm2011geometricTwo,marsden2013introduction}. We will derive the basic notions below and refer the reader these references for a broader exposition.
\begin{dfn}[Angular velocity]
    While any tangent vector can be interpreted as representing an \emph{angular velocity}\footnote{with an appropriate choice of the body-fixed frame}, it is useful to represent the {velocity} of the geodesic motion on $\SO$ as a time ($t$) dependent curve $R(t)$:
    \begin{equation}\label{eq:angvel}
        \angvel(t) = R^{\top}(t)\dR(t)\:\in\solie,
    \end{equation}
    and the corresponding \emph{angular velocity matrix} $\angvelmat:=\angvel(t)$ is skew-symmetric. 
    Since $\solie$ is isomorphic to $\R^3$, we instead store the \textbf{angular velocity vector}, $\rod\in\R^3$, corresponding to the skew–symmetric $\angvelmat$ via the \textbf{hat-operator}: $\widehat{\rod}=\angvelmat$ or $\widehat{\omega(t)}=\angvel(t)$ and equivalently $[\rod]_x=\angvelmat$.
\end{dfn}

\begin{figure}[t]
        \centering
        \includegraphics[width=\columnwidth]{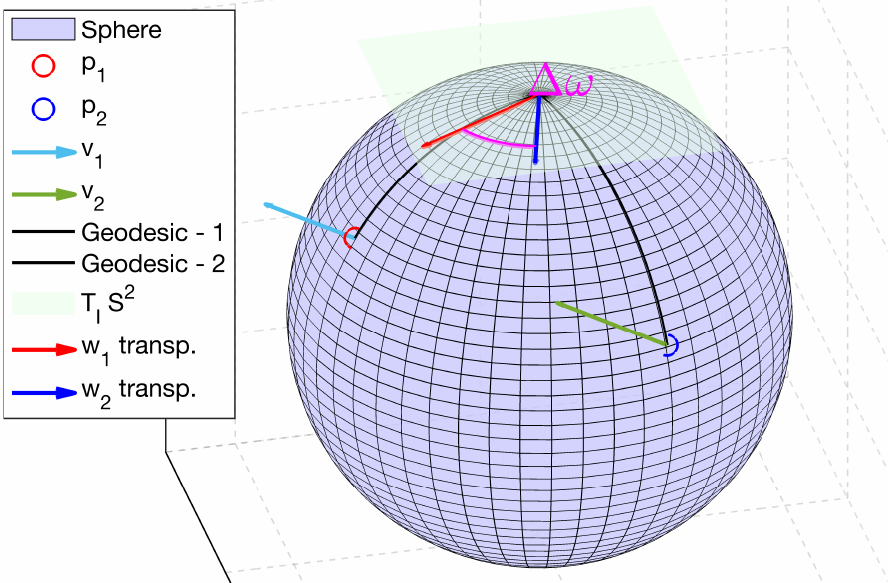}
        \caption{Illustration of angular velocity and accelaration for two points on a manifold. Given two points on the manifold ($\mathbf{p}_i$), the velocities are vectors in the tangent planes at those points. In practice, to compute an angular acceleration, they are \emph{parallel transported} to the tangent space of the identity and then compared. If the points are close enough, the variation of the tangent spaces are ignored, and the computation is carried out by directly comparing vectors. \emph{Sphere is chosen for illustration purposes whereas the manifold or articulated bodies is higher dimensional.}\vspace{-3mm}}
        \label{fig:angular_acc} 
\end{figure}
\begin{prop}[Angular acceleration]
The time-derivative of angular velocity defines the angular acceleration:
\begin{equation}
    \frac{\diff [\angvelvec_t]_x}{\diff t}:=[\drod_t]_x:=\ddot{\Rot}_t=\Rot_t\left(\angvelmat_t^2+\dot{\angvelmat_t}\right),
\end{equation}
where $\angvelmat_t^2+\dot{\angvelmat_t}$ is also skew-symmetric.
The additional $\angvelmat^2$ term accounts for the curvature of the configuration space when transforming to the inertial frame.
\end{prop}
\begin{proof}
    We start by differentiating the angular velocity in~\cref{eq:angvel} to obtain:
    \begin{align}
        \ddot{R}(t)=\dot{R}(t) \Omega(t) + R(t)\dot{\Omega}(t).
    \end{align}
    Substituting $\dot{R}(t)=R(t)\Omega(t)$ (by \cref{eq:angvel}), we obtain:
    \begin{align}
        \ddot{R}(t)&={R}(t) \Omega^2(t) + R(t)\dot{\Omega}(t)\\
        &={R}\left(t\right) \left(\Omega^2\left(t\right) + \dot{\Omega}\left(t\right) \right).
    \end{align}
    Left multiplying this with $R(t)^\top$ yields:
    \begin{align}\label{eq:accdecompose}
        R(t)^\top\ddot{R}(t)=\Omega^2\left(t\right) + \dot{\Omega}\left(t\right).
    \end{align}
    Since the square of any skew–symmetric matrix is symmetric, this is a natural decomposition of a matrix into its symmetric and skew-symmetric parts. Therefore to recover $\dot{\Omega}\left(t\right)$ as the angular acceleration, we take the skew:
    \begin{align}
    \widehat{\dot{\rod}(t)}=
    \dot{\Omega}\left(t\right) &= \skewrm\left(\Omega^2\left(t\right) + \dot{\Omega}\left(t\right)\right)\\
    &=\skewrm\left(R\left(t\right)^\top\ddot{R}\left(t\right)\right),
    \end{align}
    where the last equality follows from~\cref{eq:accdecompose}. In matrix form, $[\drod]_x=\angvelmat^2+\dot{\angvelmat}=\mathrm{\Rot^\top\ddot{\Rot}}$ where $\ddot{\Rot}$ denotes the second derivative in \emph{ambient space}.
\end{proof}

\begin{figure}[t]
    \centering
    \includegraphics[width=\linewidth]{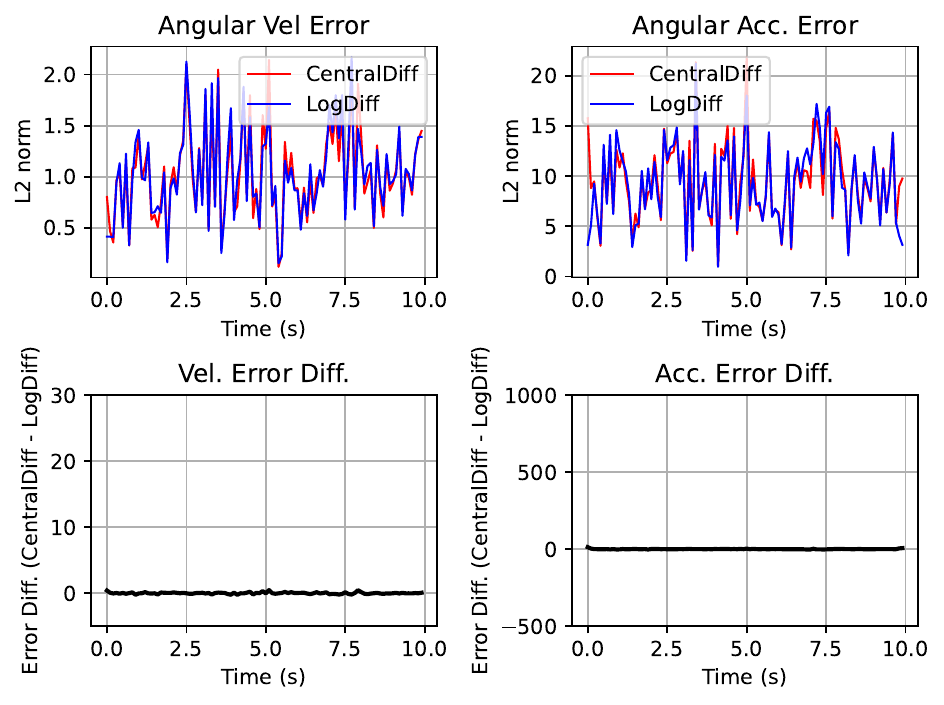}
    \caption{Comparing $\log$-central differencing to classical central differencing for angular acceleration estimation. The attained error between the two (last row) is relatively low indicating that simple central differences are good estimators of angular acceleration, given geometrically meaningful angular velocities.}
    \label{fig:angular} 
\end{figure}

\begin{remark}[On Angular Acceleration]
    In practice, we use a second-order finite differencing scheme for angular acceleration. While being more efficient than using the logarithmic maps to compare different velocities in the same tangent frame (through parallel transport as illustrated in~\cref{fig:angular_acc}), it might also be less accurate. Thus, we compare the two disceretization schemes in~\cref{fig:angular}, revealing that the two schemes are quite close in practice. This motivates us to leverage Euclidean central differencing, after estimating the angular velocities.
\end{remark}

\paragraph{Extension to power manifolds}
As articulated bodies are represented as a collection of rotations, we consider the power manifold (product of identical manifolds):
\begin{dfn}[Geodesic distances of articulated bodies]
    Endowed with the $L_p$ \emph{product metric}, the geodesic distances of articulated bodies can be written as:
    \begin{align}
        d_{\SO^K}&(\pose,\pose^\prime)\\
        &=\| d(\Rot_1,\Rot_1^\prime),d(\Rot_2,\Rot_2^\prime),\dots,d(\Rot_K,\Rot_K^\prime)\|_p,\nonumber
    \end{align}
    where $\Rot_i\in\pose\in\SO^K$ and $\Rot_i^\prime\in\pose^\prime\in\SO^K$. In this work, we use $p=1$ and $d\equiv d_{\SO}$.
\end{dfn}

\begin{dfn}[Exp / Log map on power manifold]
The natural isomorphism allows us to write its exponential map $\Exp_{\pose}:\TPose\to\SO^K$ component-wise: 
\[
\Exp_{\pose}=\left(\Exp_{\Rot_1},\Exp_{\Rot_2},\dots,\Exp_{\Rot_K}\right). 
\]
Akin to this, is the logarithmic map, $\Log_{\pose}$. 
\end{dfn}

\begin{dfn}[Riemannian gradient for an articulated pose]
 Since the tangent spaces and therefore $\Pi_{\pose}$ are replicas, the gradient of a smooth function $f:\SO^K\to\R$ w.r.t. $\pose$ is also the Cartesian product of the individual gradients:
    \begin{align}
        \grad_{\pose}{f(\pose)}  = \left(\grad_{\Rot_1}{f(\pose)},\,\dots,\,\grad_{\Rot_K}{f(\pose)}\right).
    \end{align}
\end{dfn}

\section{Algorithms, Architecture \& Details}
\paragraph{Pseudocode of RMF-Integrator}
We provide the pseudocode for integration of motion via porjection onto the learned manifold in~\cref{alg:NRMFIntegrate}.
\input{sec/alg_RMF_integrate}

\begin{figure*}[t]
    \centering
    \includegraphics[width=\linewidth]{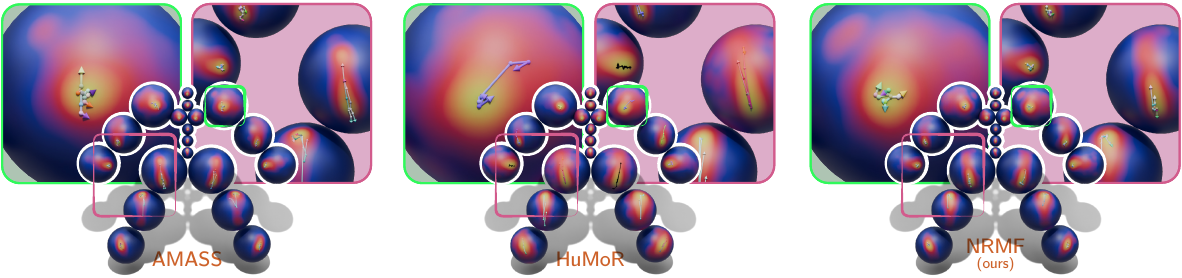}
    \caption{\textbf{Transitions and accelerations overlaid onto the pose distributions.} Four motions originating from the most common body pose in the AMASS dataset are represented as colored arrows on the surface of the spheres depicting the per-joint pose distributions comprising $p(\pose)$. Spheres here  are oriented to maximize visibility of the high-probability region of the distribution rather than showcasing the range of motions through skeletal alignment. Each transition arrow is split into different segments representing the transition between consecutive time-steps.The color of the arrows maps the three components of the angular acceleration to RGB space.}
    \label{fig:mechas} 
\end{figure*}
\paragraph{Parametric model} We employ the SMPL+H body model~\cite{smplh:SIGGRAPHASIA:2017}, consistent with the AMASS dataset~\cite{AMASS:2019} on which our work is based. Since our focus is exclusively on modeling body motion, we follow HuMoR~\cite{rempe2021humor,zhang2024rohm,shi2023phasemp} tp exclude hand joints from consideration, retaining only the 22 body joints (including the root). Incorporating hand joints into the optimization alongside body motion is straightforward, which lies outside the scope of our present study but can be a future work.

\paragraph{Test-time optimization}
We implement our optimizer and optimization pipeline with Pytorch \cite{paszke2019pytorch}. We provide full details of our HuMoR~\cite{rempe2021humor}-inspired test-time-optimization in this section. During runtime, given a sequence of 2D or 3D data observations, \eg joint detections, point clouds, described as $\obs_{0:T}$, our goal is to recover the motion sequence $\{\state_t\}_{t=0}^T$ that aligns with the observations while being \emph{likely} under the NRFM.
Our mutlti-stage optimization starts from a proper initialization using the pose prior, by minimizing the following combined objective:
\begin{equation}
    E_{I}(\roottrans,\pose,\shape) = \Ldata + \lambda_{\shape}\Lshape + \lambda_{\pose}\Lpose + \lambda_{reg}\Lreg
\end{equation}
where $\Lshape=\|\shape\|^{2}$ is the shape prior term. $\Lpose=\fpose(\pose_i)$ is our pose prior term predicting the distance to the nearest plausible pose\footnote{Note that for VPoser~\cite{pavlakos2019smplx} $\Lpose=\|\z\|^{2}_{2}$ where $z$ is the latent code.}. Moreover, we apply additional regularization terms where $\Lreg = \lambda_{\mathrm{smooth}}\Lsmooth + \lambda_{\mathrm{bl}}\Lbl$:
\begin{equation}
    \Lsmooth = \sum_{t=0}^{T}\|\joints_{t+1} - \joints_{t}\|^{2}, \quad
    \Lbl = \sum_{t=0}^{T}\|\bones_{t+1} - \bones_{t}\|^{2} \nonumber \notag
\end{equation}
where $\bones$ are the bone length of the human body. These regularization terms essentially enforces the skeleton to be consistent over time and prevent large jitters. Finally, $\Ldata$ is the task-dependent data terms and will be different when given different types of inputs. Specifically, in the task of fitting to 2D joints the objective function is written as:
\begin{equation}
    \Ldata^{\mathrm{2D}} = \lambda_{\mathrm{data}} \sum_{t=0}^{T} \sum_{j=1}^{J} \sigma_t^j \rho( \Pi(\joints_t^j) - \obs_t^j) \tag{13}
\end{equation}
where $\obs_t^j$ is the 2D joint observations with $\sigma^{j}$ as the confidence score for 2D joint observations. $\rho$ is the robust Geman-McClure function \cite{bogo2016keep,geman1987statistical} and $\Pi$ is the 2D re-projection with the 3D joints $\joints$ extracted from human meshes. When fitting 3D joints, the objective becomes:  
\begin{equation}
    \Ldata^{\mathrm{3D}} = \lambda_{\mathrm{data}} \sum_{t=0}^{T} \sum_{j=1}^{J} \| \joints_t^j - \obs_t^j \|^2 \tag{12}
\end{equation}
where the $\obs_t^j$ is the 3D joint observations. $\Ldata$ can be further extended to various observation representations such as 3D point clouds:
\begin{equation}
    \Ldata^{\mathrm{PC3D}} =
    \lambda_{\mathrm{data}} \sum_{t=0}^{T} \sum_{i=1}^{N_t} w_{\text{bs}} \min_{\Vertices_t \in V_t} \left\| \Vertices_t - \obs_t^i \right\|^2.
    \label{eq:pc3d}
\end{equation}
where $\obs$ now represents 3D human point cloud observations extracted from depth images with person segmentation masks obtained from \cite{chen2017rethinking} and $w_{\text{bs}}$ a bisquare weight \cite{beaton1974fitting} based on the Chamfer distance. 

While optimizing $E_{I}$ yields refined estimates $\state$, it cannot combat motion ambiguities since the pose prior only captures $0^{th}$-order distribution. After initialization, we factor in our transition and acceleration priors as $\Ltransition$ and $\Lacceleration$:
\begin{equation}
    \Ltransition := \fvel(\posevel_i) \quad \Lacceleration := \facc(\poseacc_i)
\end{equation}

 \paragraph{Disentangle and model global motion representation} For our motion state representation $\state$, we explicitly disentangle global motion (relative to the current state) $\roottrans \in \mathbb{R}^{3}$ from the local pose configuration for each state. Specifically, we achieve this by multiplying the inverse root transform with each pose, mapping all poses into a local coordinate frame aligned to a canonical facing direction. This canonicalization simplifies the learning process by removing unnecessary global variation. Inspired by prior works \cite{he2022nemf,rempe2021humor}, we predict $\{\roottrans\ , c_t^i\}, $ through the sub-module network integrated within the transition prior framework based on the same rich contextual input following \cite{rempe2021humor}. During optimization, it is constrained by data terms during the optimization. Additionally, we expand $\Lreg$ to account also for floor contact physics constraint terms (contact  loss $\Lcontact$ and contact height loss $\Lch$):
\begin{equation}
        \Lcontact = \sum_{t=1}^{T} \sum_{i=1}^{N_{contact}} \lambda_{\text{cj}} c_t^i \left\| \joint_t^i - \joint_{t-1}^i \right\|^2 + \lambda_{\text{cv}} c_t^i \left\| \posevel_{t}^{i} \right\|^2
\end{equation}
\begin{equation}
    \Lch = \sum_{t=1}^{T} \sum_{i=1}^{N_{contact}} \lambda_{\text{ch}} c_t^i \max (|\joint_{z,t}^i| - \delta, 0),
\end{equation}
where $c_t^i$ is the contact probability predicted by the transition prior network specifically for joint $i$ at timestep $t$, and the contact height loss will further constrain the z-component of $\joint^{i}$. The updated objective function then becomes:
\begin{equation}
\begin{aligned}
    E_{I}^\prime(\roottrans,\pose,\shape) = \Ldata 
    + \lambda_{\shape}\Lshape 
    + \lambda_{\pose}\Lpose \\
    + \lambda_\mathrm{reg}\Lreg 
    + \lambda_{\posevel}\Ltransition 
    + \lambda_{\poseacc}\Lacceleration
\end{aligned}
\end{equation}
Moreover, we apply the rollout in \cref{alg:NRMFIntegrate} within each iteration to update the motion state $\state$ for drift correction and rebuild the consistent state. During the test-time optimization of motion estimation, we balance the loss terms and prior with the following coefficients: $\lambda_\mathrm{2d}=1e^{-3},\lambda_\mathrm{smooth}=10, \lambda_{\shape}=8e^{-2}, \lambda_{\pose}=8e^{-2},\lambda_{\posevel}=1.0,\lambda_{\posevel}=5e^{-2} $. The latter terms influence the trust in each prior.

\paragraph{Training data generation}
We now provide more details of our training process, including the data preparation. For training dataset, we follow \cite{tiwari22posendf,he24nrdf,lu2023dposer} to use the same split and clip the motion sequence to extract the middle 80\% and downsample the dataset to 30 fps. We then sample the negative training samples with an adopted multi-step mechanism for querying the nearest neighbours using FAISS \cite{johnson2019billion} to speed up the NN process. To accelerate nearest-neighbor (NN) search, we adopt a two-stage retrieval scheme similar to \cite{tiwari22posendf,he24nrdf}. In the first stage, we use FAISS \cite{johnson2019billion} to retrieve $k'$ coarse candidates for each noisy transition or acceleration. In the second stage, we refine these candidates by computing the distance and selecting the closest $k$ neighbors. In our setup, we set $k' = 1000$ and $k = 1$. Unlike Pose-NDF, which averages the distances to the top 5 nearest neighbors, we determine the final match using only the single closest neighbor. To generate negative samples, we adopt a mixed sampling strategy with fixed ratios for three types of negatives: distance-targeted half-Gaussian distribution based perturbations (60\%) following the scheme in \cite{he24nrdf}, random-swap candidates (30\%), and fully random candidates (10\%).  
Specifically, for the first category, we perturb the ground-truth transition or acceleration by applying a small random rotation sampled from a zero-mean Gaussian distribution, thereby introducing controlled manifold noise.  
For the second category, we randomly select transitions from the clean motion dataset without ensuring semantic proximity to the query.  
For the third category, we generate fully random quaternions and normalize them to unit length.  
All negative candidate are concatenated and passed through a single FAISS-based nearest neighbor search to compute their distances to the training set.

\paragraph{Training details} In our experiments, we introduce our baselines HuMoR \cite{rempe2021humor}, RoHM \cite{zhang2024rohm}, and MDM \cite{tevet2023human} which were trained on the same split as ours for a fair comparison. Specifically, the original MDM essentially requires clean motion because it would replace denoised joints
 with visible input joints at each timestep of denoising and was not trained on AMASS. To make it comparable on AMASS, we train MDM with the same split and with quaternion as the rotation representation instead of the original $hml_{vec}$, which is a high-dimensional composition vector of motion states. For PhaseMP \cite{shi2023phasemp}, we use the official code for evaluation and training. Finally, our prior modules (T-NRDF and A-NRDF) are pre-trained independently in the first stage and jointly optimized in the second stage with learning rate of 3e-5. For training the global motion prediction layers, we follow a simple scheme to train it jointly with the network in our experiments. We choose the checkpoints by selecting the best-performing model on validation set.

\begin{figure*}[t]
        \centering
        \includegraphics[width=\linewidth]{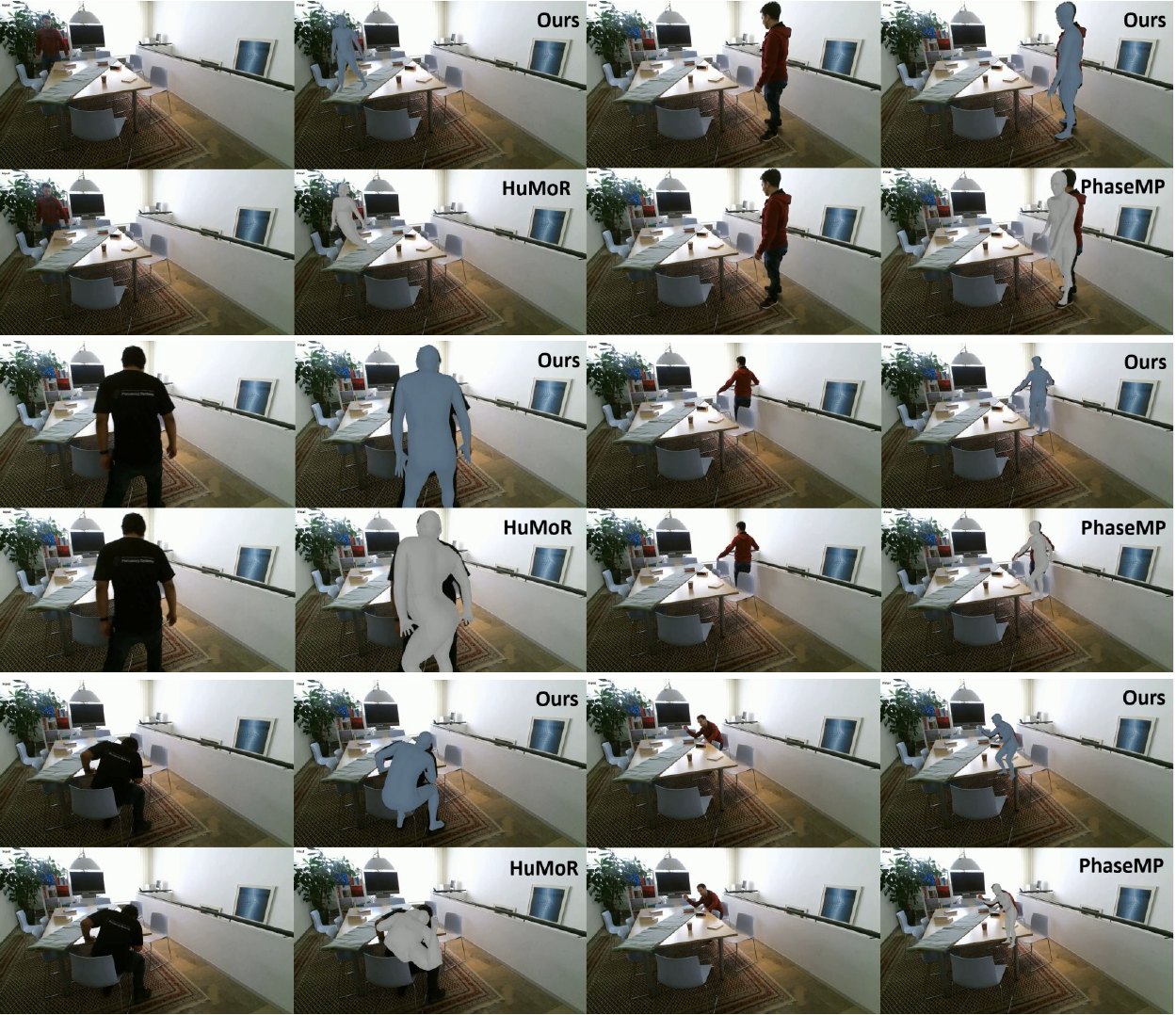}
        \caption{ {Qualitative results on in-the-wild fitting on PROX \cite{hassan2019prox} dataset. We compare our method agains \cite{zhang2024rohm,rempe2021humor,shi2023phasemp}, which demonstrates the robustness of our method under challenging occlusion.}\vspace{-4mm}}
        \label{fig:PROX} 
\end{figure*}

\begin{figure*}[t]
        \centering
        \includegraphics[width=\linewidth]{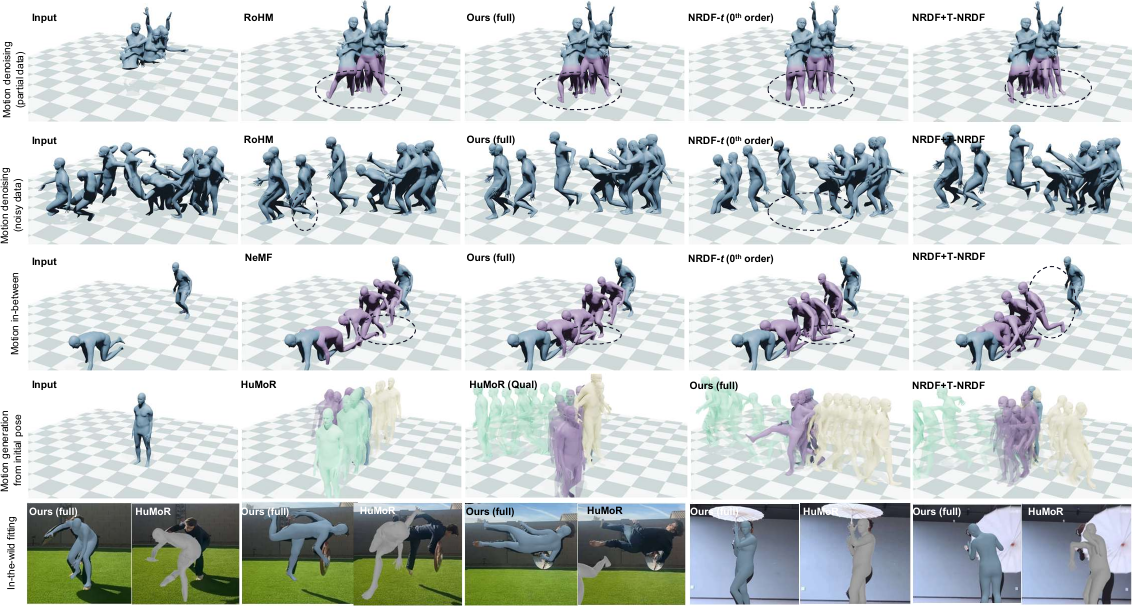}
        \caption{{Qualitative results on downstream applications.} The \textcolor{steelblue}{\textbf{mesh and body parts}} are the observation inputs and we show the \textcolor{SoftLavender}{\textbf{output or optimization results}}. For in-the-wild fitting, we conduct experiments on both in-door datasets (bottom row) and online videos (upper row).\vspace{-4mm}}
        \label{fig:qua_supp_1} 
\end{figure*}

\begin{figure*}[t]
        \centering
        \includegraphics[width=\linewidth]{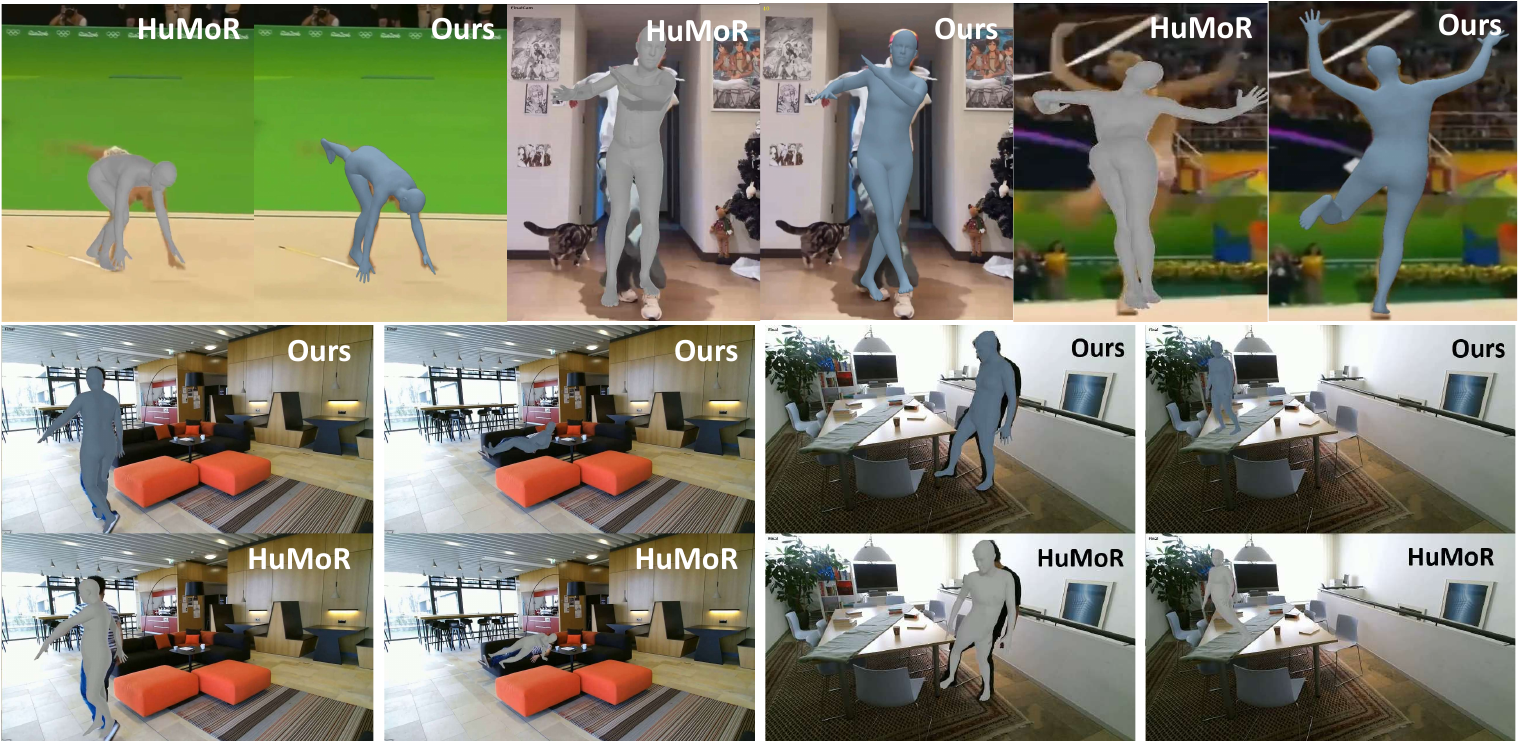}
        \caption{{In-the-wild motion estimation. We show qualitative comparisons on real-world videos with challenging conditions such as occlusion, background clutter. Our method recovers more plausible, consistent, and anatomically accurate motions.} \vspace{-4mm}}
        \label{fig:qua_supp_2} 
\end{figure*}

\paragraph{Details on joint distribution visualizations}
    Although poses are defined in the product manifold of $\SO$ ($\pose \in \SO^{N_{J}-1}$), for the purpose of visualization, we separately analyze each joint angle space. We observe that joint rotations can be plotted onto the surface of a sphere by rotating skeletal bones in their rest position. Precisely, given a unit vector representing the direction of the axis between two consecutive joints in the kinematic tree, we rotate it using the Rodrigues' rotation formula and obtain a rotated unit vector. Repeating this procedure for a set of joint poses, we obtain a set of points on a sphere. We thus fit these points using a spherical kernel density estimation~\cite{handley2020sphericalkde}, which replaces the traditional Gaussian function used in kernel density estimation with the Von Mises-Fisher distribution. We repeat the same procedure for both $\pose$ and $\posevel$, as $\posevel$ is also a relative angle determing the transition between two poses. 

    All spherical kernel density estimations are performed on a set of $10,000$ rotations using a fixed bandwidth $h = 0.08$. The log-probability density estimates computed at the vertices of each sphere are subject to a power normalization with exponent $\gamma$. For ease of visualization, we set $\gamma=60$ for the poses of AMASS, $\gamma=50$ for NRMF, and $\gamma=20$ for NRDF, HuMoR, and all the transition densities. 

    Poses and transitions are randomly selected from AMASS, and generated with all the other methods. For NRMF, we select $10,000$ samples from $195$ $150$-frames-long motion sequences generated starting from initial poses of AMASS motions. 

    In addition to the evaluation of the pose $p(\pose)$ and transition $p(\posevel)$ distributions, we also explicitly visualize a selected subsets of motion sequences and their full states $\state_t$. We choose the most common pose in AMASS, denoted by $\pose_c$, as the starting point for all trajectories and methods. We then perform a $k$-nearest neighbor ($k$NN) search using $d_{\SO}^{N_J}(\pose_c, \pose_i)$ to geodesically measure distances and correctly identify the $150$ closest poses to $\pose_c$. These poses can occur at any timestep of a motion sequence as long as there are enough timesteps to complete a $20$ frames motion. To promote diversity among the motions to represent, we also perform a  $d_{\SO}^{N_J}$-based farthest point sampling with respect to $\pose_c$ and all the poses at $t+20$ extracted from the motion sequences identified with $k$NN. This allows us to identify the $4$ trajectories of $20$ timesteps originating as close as possible to $\pose_c$ and terminating the farthest apart from each other --and their origin. 
    
    The trajectories are geodesically traced onto the spheres using FlipOut, the edge flipping algorithm introduced in \cite{sharp2020flipout} to find shortest geodesics on polyhedral surfaces. Note that we rely on a discrete method for geodesic computations because the spheres are effectively discretized into ico-spheres for visualization purposes. We break the trajectory in $3$ segments each containing the same number of timesteps. Therefore, longer segments equate to faster transitions. Accelerations are finally represented mapping the three components of $\poseacc$ to the RGB space.  

    As it can be observed in \cref{fig:all_distribs}, HuMoR tends to overestimate the range of motion of each joint, often enabling poses which are never or rarely represented in the training data. NRDF significantly mitigates this issue, closely mimicking the behavior of AMASS on most joints. Nevertheless, NRDF appears the best at capturing the true pose distribution. A similar trend is observable for transitions, which not only appear slightly improved in terms of distributions in \cref{fig:all_distribs}, but also in terms of individual examples reported in \cref{fig:mechas}. In fact, the arrow orientations of NRMF more closely align with those of AMASS. Most notably, when comparing accelerations, we note that while NRMF can properly capture their variability, HuMoR seems to erroneously produce transitions with almost-fixed angular accelerations. These results prove once again the superiority of our method compared to the state of the art as well as the necessity of adopting a motion model.

\section{Additional Evaluations}
In this section we provide more qualitative results for our motion estimation. As shown in \cref{fig:qua_supp_1} and \cref{fig:qua_supp_2}, we present additional qualitative comparisons on motion estimation and in-betweening. Our method consistently outperforms existing approaches such as HuMoR \cite{rempe2021humor}, RoHM \cite{zhang2024rohm}, and PhaseMP \cite{shi2023phasemp}, producing more realistic and physically plausible motions. In scenarios with partial or noisy observations, our predictions exhibit improved temporal consistency and anatomically accurate transitions, especially in challenging regions like occluded body parts. For motion in-betweening and motion generation, NRMF generates smooth and coherent transitions between sparse keyframes, capturing natural dynamics and avoiding artifacts or discontinuities often seen in baseline methods. These results further highlight the benefit of incorporating higher-order priors and respecting the underlying geometry of articulated motion.

\paragraph{Motion In-betweening} 
\begin{table}[t]
\centering
\caption{\textbf{Sparse keyframe in-betweening}. We report FID$_{m}$ and Acc Err for different sequence lengths. The best and second-best results are shown in \textbf{bold} and \underline{underlined}, respectively.}
\label{tab:inbetweening}
\resizebox{\linewidth}{!}{%
\begin{tabular}{l|cccc|cccc}
\toprule
& \multicolumn{4}{c|}{\textbf{FID$_{m}$} $\downarrow$} & \multicolumn{4}{c}{\textbf{Acc Err ($m/s^{2}$)} $\downarrow$} \\
Method & End points & 10\% & 20\% & 30\% & End points & 10\% & 20\% & 30\% \\
\midrule
SLERP \cite{shoemake1985slerp} & 3.418 & 0.321 & \underline{0.031} & \underline{0.024} & 19.357 & 17.226 & 9.207 & 5.371\\
HuMoR \cite{rempe2021humor} & 2.572 & 0.695 & 0.106 & 0.078 & 13.356 & 11.579 & 6.601 & 3.378 \\
NeMF~\cite{he2022nemf} & \underline{2.044} & 0.315 & 0.081 & 0.063 & 12.864  & 10.194 & 6.535 & 3.563 \\
\hline
Ours (SLERP) & 2.238 & \underline{0.308} & \textbf{0.029} & \textbf{0.019} & \underline{10.767} & \underline{6.478} & \underline{3.768} & \underline{2.319} \\
\rowcolor{gray!15} Ours (NeMF) & \textbf{1.891} & \textbf{0.301} & 0.045 & 0.032 & \textbf{7.135} & \textbf{5.245} & \textbf{3.394} & \textbf{1.983}\\
\bottomrule
\end{tabular}
}
\end{table}
To further evaluate {\name}'s performance in reconstructing plausible motion under sparse supervision, we follow the sparse keyframe in-betweening protocol introduced in \cite{he2022nemf}. Specifically, we provide a small subset of keyframes as observations, either the two endpoints or $20\%$, $35\%$, and $50\%$ of the total frames, and evaluate the model’s ability to recover the full motion sequence. As shown in Table~\ref{tab:inbetweening}, {\name} consistently improves the baseline methods and achieves lower FID scores, indicating higher visual fidelity of the generated motions. In addition, we report the Acc Err as a measure of temporal smoothness and physical realism. Our method outperforms both NeMF across all sparsity levels, achieving the best trade-off between accuracy and dynamical consistency. These results highlight {\name}’s ability to robustly interpolate missing frames by leveraging high-order dynamics.

\paragraph{Computational Cost} Excluding the time to obtain initial observations, our optimizer will take approximately 7min. to process a video of 10sec. at 30 fps on an NVIDIA A100 GPU. We also report the processing time for a 30 fps 10-second RGB video. As seen, our NRMF reduces the turn-around for 10 times, which is an order of magnitude lighter than the widely used HuMoR.
\begin{table}[ht!]
  \centering
  \footnotesize
  \vspace{-3mm}
  \setlength{\tabcolsep}{3.5pt}
  \label{tab:runtime-horizontal}
  \begin{tabular}{l|c|ccc|c}
          & HuMoR & Stage I & Stage II & Stage III
          & {\textbf{Total (Ours)}} \\
    \midrule
    Runtime (min.) &
      61.56 & 0.62 & 2.18 & 3.23 & 6.03 \\
  \end{tabular}
  \vspace{-3mm}
\end{table}

%% file: sec/alg_RMF_integrate.tex
\begin{algorithm}[t]
\caption{RMF Integrator}
    \begin{algorithmic}[1]
    \Require An initial pose $\pose_0$ and velocity $\posevel_0$, a possibly noisy acceleration sequence $\{\poseacc_t\}_t$ and all components of the trained network $\fall$ (and hence the projectors $\Pi$),
    \Ensure Plausible motion $\{\X_t\}_t$
        \For{$t\gets 1,\dots, T$} \Comment{Euler integrate of velocities}
            \State $\posevel_t \gets \posevel_{t-1}+\lambda_t\poseacc_{t-1}$ 
            \State $\posevel_t \gets \VelProj(\posevel_t)$
        \EndFor
        \For{$t\gets 1,\dots, T$} \Comment{Update poses}
            \State $\pose_{t} \gets \Exp_{\pose_{t-1}}\bigl(\alpha_t\, [{\posevel}_{t-1}]_x\bigr)$
            \State $\pose_{t} \gets \PoseProj(\pose_{t})$
        \EndFor
    \State Compose $\{\X_t\}_t$ from individual estimates $\{\poseacc_t, \posevel_t, \pose_t\}_t$
    \end{algorithmic}
    \label{alg:NRMFIntegrate} 
\end{algorithm}